\documentclass[sigconf]{acmart}
\AtBeginDocument{%
  }

\copyrightyear{2025}
\acmYear{2025}
\setcopyright{acmlicensed}
\acmConference[WWW '25]{Proceedings of the ACM Web Conference 2025}{April 28-May 2, 2025}{Sydney, NSW, Australia}
\acmBooktitle{Proceedings of the ACM Web Conference 2025 (WWW '25), April 28-May 2, 2025, Sydney, NSW, Australia}
\acmDOI{10.1145/3696410.3714962}
\acmISBN{979-8-4007-1274-6/25/04}

\usepackage{xcolor}
\usepackage{colortbl}
\usepackage{float}
\usepackage{xspace}
\usepackage{amsmath}
\usepackage[capitalize,noabbrev]{cleveref}
\usepackage{graphicx}
\usepackage{pifont}
\usepackage{enumitem}
\usepackage{multirow}
\usepackage{multicol}
\usepackage{tablefootnote}
\usepackage[most]{tcolorbox}
\usepackage{wrapfig}
\usepackage{bbding}
\usepackage{subcaption}
\usepackage{adjustbox}
\usepackage[normalem]{ulem}
\usepackage{subcaption}
\usepackage[ruled,linesnumbered]{algorithm2e}
\usepackage{listings}

\newcommand{\rmnum}[1]{\romannumeral #1}
\newcommand{\Rmnum}[1]{\uppercase\expandafter{\romannumeral #1}}
\definecolor{dg}{rgb}{0.0, 0.6, 0.0}
\definecolor{lb}{rgb}{0.9, 0.95, 1.0}
\definecolor{purple}{rgb}{0.3, 0.1, 0.4}
\newcommand{\up}{\color{purple}{$\uparrow$}\xspace}

\begin{document}

\title{SheetAgent: Towards a Generalist Agent for Spreadsheet Reasoning and Manipulation via Large Language Models}


\author{Yibin Chen}
\authornote{Both authors contributed equally to this research.}
\email{yibin\_chen@tju.edu.cn}
\affiliation{%
  \institution{School of New Media and Communication, Tianjin University}
  \city{Tianjin}
  \country{China}}

\author{Yifu Yuan}
\authornotemark[1]
\email{yuanyf@tju.edu.cn}
\affiliation{%
  \institution{College of Intelligence and Computing, Tianjin University}
  \city{Tianjin}
  \country{China}}

\author{Zeyu Zhang}
\email{3020208104@tju.edu.cn}
\affiliation{%
  \institution{College of Intelligence and Computing, Tianjin University}
  \city{Tianjin}
  \country{China}}

\author{Yan Zheng}
\authornote{Corresponding author: Yan Zheng (yanzheng@tju.edu.cn)}
\email{yanzheng@tju.edu.cn}
\affiliation{%
  \institution{School of New Media and Communication, Tianjin University}
  \city{Tianjin}
  \country{China}}

\author{Jinyi Liu}
\email{jyliu@tju.edu.cn}
\affiliation{%
  \institution{College of Intelligence and Computing, Tianjin University}
  \city{Tianjin}
  \country{China}}

\author{Fei Ni}
\email{fei\_ni@tju.edu.cn}
\affiliation{%
  \institution{College of Intelligence and Computing, Tianjin University}
  \city{Tianjin}
  \country{China}}

\author{Jianye Hao}
\email{jianye.hao@tju.edu.cn}
\affiliation{%
  \institution{College of Intelligence and Computing, Tianjin University}
  \city{Tianjin}
  \country{China}}

\author{Hangyu Mao}
\email{hy.mao@pku.edu.cn}
\affiliation{%
  \institution{Independent Researcher}
  \city{Beijing}
  \country{China}}

\author{Fuzheng Zhang}
\email{zhfzhkris@outlook.com}
\affiliation{%
  \institution{Independent Researcher}
  \city{Beijing}
  \country{China}}

\renewcommand{\shortauthors}{Yibin Chen et al.}

\begin{abstract}
    Spreadsheets are ubiquitous across the World Wide Web, playing a critical role in enhancing work efficiency across various domains. Large language model (LLM) has been recently attempted for automatic spreadsheet manipulation but has not yet been investigated in complicated and realistic tasks where reasoning challenges exist (e.g., long horizon manipulation with multi-step reasoning and ambiguous requirements). To bridge the gap with the real-world requirements, we introduce \textbf{SheetRM}, a benchmark featuring long-horizon and multi-category tasks with reasoning-dependent manipulation caused by real-life challenges. To mitigate the above challenges, we further propose \textbf{SheetAgent}, a novel autonomous agent that utilizes the power of LLMs. SheetAgent consists of three collaborative modules: \textit{Planner}, \textit{Informer}, and \textit{Retriever}, achieving both advanced reasoning and accurate manipulation over spreadsheets without human interaction through iterative task reasoning and reflection. Extensive experiments demonstrate that SheetAgent delivers 20--40\% pass rate improvements on multiple benchmarks over baselines, achieving enhanced precision in spreadsheet manipulation and demonstrating superior table reasoning abilities. More details and visualizations are available at the \href{https://sheetagent.github.io/}{project website}. The datasets and source code are available at \url{https://anonymous.4open.science/r/SheetAgent}.
\end{abstract}

\ccsdesc[500]{Computing methodologies~Natural language processing}
\ccsdesc[500]{Information systems~Decision support systems}

\keywords{Agents, Large Language Models, Benchmark, Spreadsheet Reasoning and Manipulation}


\maketitle

\begin{figure*}[!t]
\centering  
\includegraphics[width=0.95\textwidth]{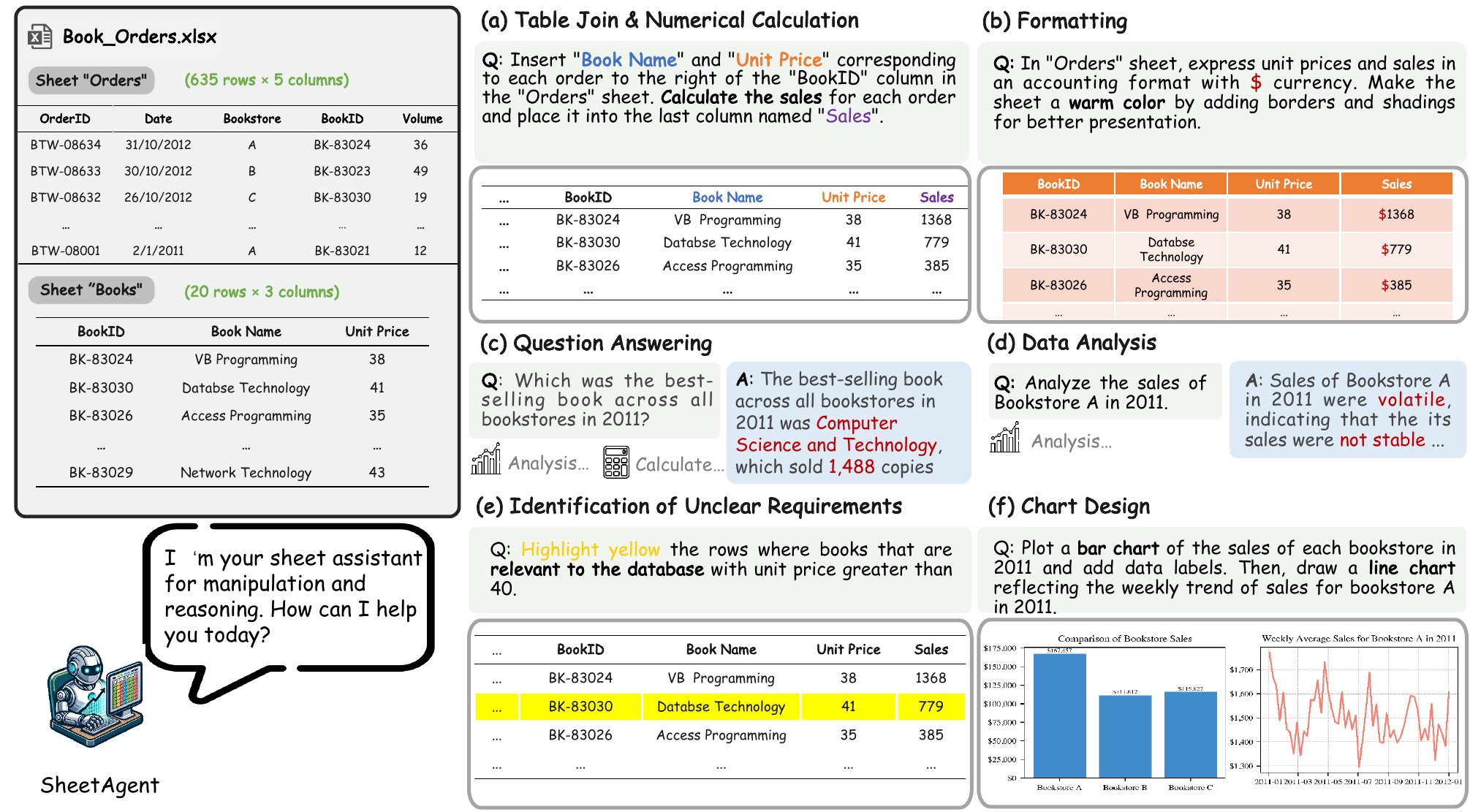}
\caption{\textbf{SheetAgent can handle diverse spreadsheet reasoning and manipulation tasks automatically.} Given a large-scale spreadsheet with multiple sheets, SheetAgent shows its proficiency in visualization (\textbf{f}), achieves accurate manipulation on long horizon tasks (\textbf{a, b}) with consistent reasoning capabilities (\textbf{c, d}), even faced with the challenges like unclear requirements (\textbf{e}).}
\label{fig:overview}
\end{figure*}

\section{Introduction}
\label{sec:intro}
Tabular data plays a crucial role in domains such as scientific research, finance, and marketing, where it is predominantly handled using spreadsheet systems. These systems, such as Google Sheets and Microsoft Excel, are useful for tasks including numerical calculations, data analysis, and visualisation \cite{kandel2012enterprise, hasan2020current, edeling2021marketing}. However, processing these affairs often involves a significant amount of repetitive labor and consultation \cite{gulwani2011automating, chen2021spreadsheetcoder}. Recent work \cite{chen2021spreadsheetcoder, li2024sheetcopilot} has explored the automation of simple spreadsheet manipulation tasks. For example, \textit{Highlight rows with sales volume greater than 40} in the ``Orders'' sheet shown in \cref{fig:overview}. This task can be accomplished through simple queries and formatting. However, they fail to consider the more complex and realistic tasks that encompass more than simple first-order logic. For instance, \textit{Highlight rows of database-related books with sales volume greater than 40}. The difficulty of this instruction lies in identifying books related to the database, which cannot be achieved simply by understanding the semantics of column names, but rather by perceiving the specific content (what books are in the spreadsheet in this case) of the spreadsheet. Such scenarios are common because a complete spreadsheet\footnote{A spreadsheet is a collection of sheets that are organized into a document. A table represents a structured arrangement of data in rows and columns. Each sheet within the spreadsheet contains a table.} task may demand a multi-step reasoning process in conjunction with multiple sheets, and the user may not precisely define the required operations or ambiguously interpret the task instruction. Consequently, there is an urgent need for a new method to automate these tasks.

Designing such a method demands a combination of sophisticated sheet-based reasoning and manipulation capabilities. Previous work \cite{gulwani2014nlyze, payan2023instructexcel, li2024sheetcopilot} has focused on precise spreadsheet manipulation while neglecting reasoning, limiting them to tasks with clear expressions and one-step reasoning. The emergence of large language models (LLMs) like GPTs \cite{radford2018improving, radford2019language, brown2020language} enables the integration of reasoning and manipulation capabilities. Extensive research \cite{chen2023large, ye2023large, jiang2023structgpt} has shown that LLMs can reason over tables, handling tasks such as table question answering and fact verification. Given this context, we are motivated to explore the question: \textit{Can we build a versatile agent adept at handling complex spreadsheet manipulation tasks with challenging reasoning factors using LLMs?} Designing an automated system is crucial \cite{zheng2019wuji,yuanuni,ni2024generate,liu2024enhancing}. However, crafting such an automatic agent involves several challenges: (1) \textbf{Dynamic Changes of Spreadsheet}: Complex tasks often involve multiple operations, resulting in dynamic spreadsheet content changes. Continuously feeding the entire spreadsheet into LLMs before each operation is impractical due to token limits and potential hallucination \cite{cheng2022binding, ye2023large}. (2) \textbf{Limited Table Understanding}: LLMs are predominantly trained in natural languages and show limited understanding of tables \cite{li2023table}. (3) \textbf{Lack of Benchmark}: There is an absence of a complicated benchmark demanding accurate reasoning and precise manipulation over spreadsheets. SheetCopilot \cite{li2024sheetcopilot} presents a benchmark for evaluating LLM performance in controlling spreadsheets. However, it simplifies real-world requirements, ignoring challenges like multi-step reasoning and long-horizon operations.

To address the dataset gap, we first introduce \textbf{SheetRM} (\textbf{Spread\-sheet} \textbf{R}easoning and \textbf{M}anipulation Benchmark), a benchmark for developing and evaluating LLM-based agents for precise spreadsheet manipulation and advanced reasoning capabilities. Each task in SheetRM involves multiple subtasks that relies on reasoning abilities, derived from real-world Excel exam datasets. Moreover, it enables automatic evaluation with various metrics. We further present \textbf{SheetAgent}, a generalist agent for sheet manipulation and reasoning using LLMs. SheetAgent mainly consists of three components: the Planner, Informer, and Retriever. The Planner translates conceptual understandings into proficient code generation to manipulate spreadsheets. The Informer parses task demands and produces high-quality, task-specific SQL queries to understand the spreadsheet without needing to read the entire table, despite its dynamic changes. The Retriever retrieves instructive examples to improve the robustness of solutions. We demonstrate that SheetAgent significantly outperforms other state-of-the-art baselines in diverse benchmarks.  Our contributions are three-fold:

\begin{itemize}[leftmargin=*, ]
    \item We introduce SheetRM, a benchmark for developing and evaluating LLM-based agents to manipulate spreadsheets with advanced reasoning abilities. It includes more challenging tasks that reflect real-world requests and supports automatic evaluation with various metrics.
    \item We develop a versatile LLM-based agent SheetAgent, combining sheet manipulation and reasoning abilities to boost multifaceted interaction between humans and spreadsheets.
    \item Experimental results show that SheetAgent can be combined with any LLMs backbone and SheetAgent outperforms baselines across multiple benchmarks, achieving a 20--40\% improvement in various metrics. These results highlight SheetAgent's exceptional capabilities in spreadsheet manipulation and table reasoning.
\end{itemize}

\begin{figure*}[!t]
\centering
\includegraphics[width=0.98\textwidth]{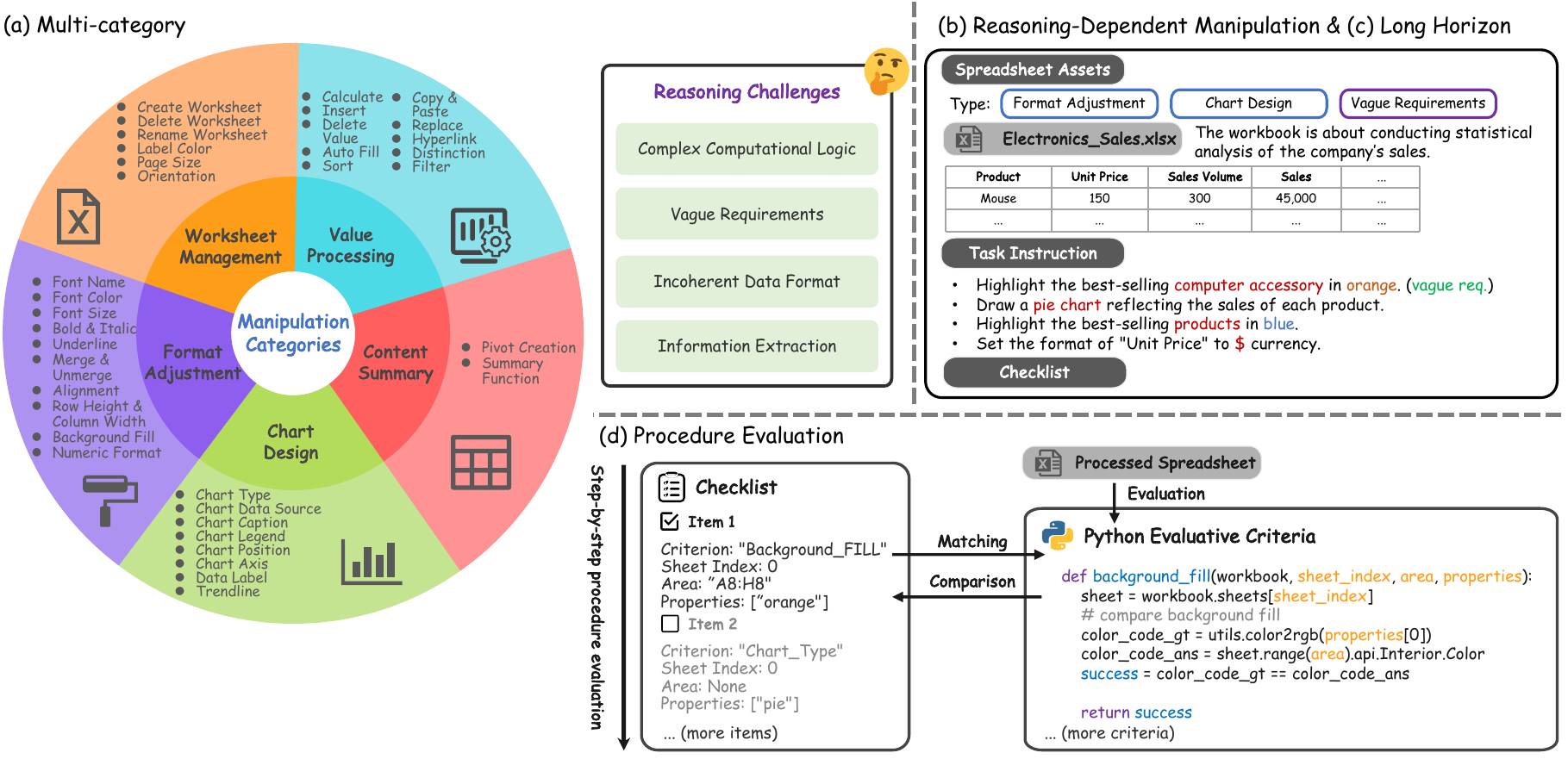}
\caption{Overview and features of SheetRM. \textbf{(a) Multi-category:} SheetRM contain real-life tasks for multiple types of manipulation categories and reasoning challenges. Each task includes an examination of both manipulation and reasoning abilities. \textbf{(b\&c) Long horizon and reasoning-dependent Manipulation}: An example task including three parts. Spreadsheet assets contain sheet data and one-sentence description with category of tasks. Then task instruction provides the requirements for the execution of the long horizon tasks. Checklist is designed for procedure evaluation. \textbf{(d) Procedure evaluation}: SheetRM automatically evaluates each task step-by-step via corresponding checklist and evaluative criterion to achieve procedure evaluation.}
\label{fig:benchmark}
\end{figure*}

\section{SheetRM Benchmark}
\label{sec:benchmark}
Unlike existing datasets \cite{gulwani2012spreadsheet,payan2023instructexcel, li2024sheetcopilot} primarily designed for more precise spreadsheet manipulation, our goal is to construct a more realistic dataset, where tasks contain challenges such as complicated multi-step reasoning and vague requirements, to narrow the gap between simulation and real-world scenarios.
To achieve this, we begin by sourcing spreadsheets from real-life Excel exam datasets. 
We collate a diverse set of spreadsheet operations commonly used in realistic scenarios and analyze the challenges faced when addressing spreadsheet tasks in practical settings. In brief, our SheetRM dataset is featured by the following elements, as outlined in \cref{fig:benchmark}(a)-(d):
\begin{itemize}[leftmargin=*]
    \item \textbf{Multi category:} We summarize and collect 5 broad types and 36 subtypes of manipulation categories with corresponding 4 reasoning challenges. Each task includes an examination of both manipulation and reasoning abilities.
    \item \textbf{Reasoning-dependent manipulation:} Tasks include operations with multi-step reasoning over spreadsheets.
    \item \textbf{Long horizon:} Various subtasks constitute a complete task, which brings to agents the challenge of dynamic changes in spreadsheets.
    \item \textbf{Procedure evaluation:} We build an automated program evaluation approach for SheetRM that not only automates the determination of whether the full task is completed but also detects the completion of individual subtasks.
\end{itemize}

\subsection{Task Schema}
\label{sec:task_schema}
Each task is defined by the following three parts. See \cref{fig:benchmark} (upper right) for a visual demonstration. 

\noindent\textbf{Spreadsheet Assets.} Each task consists of a spreadsheet as well as multiple sheets. We summarize the contents of the spreadsheet in a one-sentence natural language overview as context, as shown in \cref{tab:sheet_desc}, aiming to stimulate the internal knowledge of LLMs.

\noindent\textbf{Task Instruction.} A task instruction outlines the overall requirements expressed in natural language. Completing a task instruction requires a series of operations on the target spreadsheet.

\noindent\textbf{Checklist.} 
A task is paired with a checklist designed to evaluate its completion. Each item in the checklist corresponds to the evaluation of a fine-grained operation with tailored criteria. The automatic evaluation will be discussed in detail in \cref{sec:automatic evaluation}.

\subsection{Dataset Construction}
\label{sec:data_construction}
We gather and refine publicly available spreadsheets through a selection and cleaning process. Tasks are generated with both human and GPT-4 annotation. All the tasks are attached with verified answers, which enables model-free evaluation. The statistics of our curated dataset are shown in \cref{tab:basic_stat}. Compared to the SheetCopilot benchmark, our SheetRM has a more granular and reasonable categorisation, holds more tasks with longer horizon and includes reasoning challenges. See \cref{app:sheetrm_detail} for a detailed comparison and more statistics.

\begin{table}
\centering
\caption{Basic statistics of SheetRM.}
\label{tab:basic_stat}
\begin{tabular}{lc}
\toprule
\textbf{Item}                 & \textbf{Count}  \\ \midrule
\# Sheets                   & 137     \\
\# Average Rows per File    & 300.82 \\
\# Average Columns per File & 26.23  \\
\# Tasks        & 317    \\
\# Subtasks                 & 1625   \\ \bottomrule
\end{tabular}
\end{table}


\noindent\textbf{Spreadsheet Files Collection.} We began by gathering spreadsheets from a public examination question bank, carefully excluding files that were protected, corrupted, or inaccessible. To address privacy concerns, we modified sensitive information by adding noise to age data and anonymizing names, such as bookstores. Our selection aimed to cover multiple domains, ensuring diversity. Each file typically contains at least two sheets with a minimum of 20 rows and 5 columns. External dependencies were translated into natural language or embedded as feasible. Ultimately, we selected 41 spreadsheets with 137 sheets, averaging 300.82 rows and 26.23 columns per spreadsheet. This comprehensive collection provides a rich dataset for in-depth analysis. For additional details on our collection process, refer to \cref{app:collection}.

\noindent\textbf{Task Generation.} We begin by referring to websites about spreadsheet software skills and consult corporate staff about commonly used spreadsheet operations in their work. As shown in \cref{fig:benchmark}, we conclude five coarse operation categories and their fine-grained specific operations for manipulation. Drawing insights from common table reasoning datasets like WikiTableQuestions, FeTaQA, and TabFact, we summarize four challenges in the process of sheet reasoning: (1) complex computation logic, (2) vague requirements, (3) incoherent data format and (4) information extraction. We detail these challenges in the \cref{app:challenges}. Then, we instruct GPT-4 to propose \emph{realistic} tasks that mimic user requests adhering to four guidelines: the tasks should only involve predefined operations, cover diverse manipulation categories, exhibit a long-horizon nature by encompassing multiple subtasks, and incorporate at least one subtask that presents the specified reasoning challenges. This process yields a compilation of 2316 subtasks. We eliminate semantically redundant entries for identical files to maintain uniqueness. To guarantee quality, our internal annotators manually validate subtasks using programming and specialized software such as Excel. Certain unreasonable subtasks are excluded throughout this process. By combining these subtasks for different spreadsheets considering horizon and complexity, we ultimately assemble 317 task instructions, encompassing a total of 1625 subtasks.

\begin{figure*}[!t]
\centering
\includegraphics[width=\textwidth]{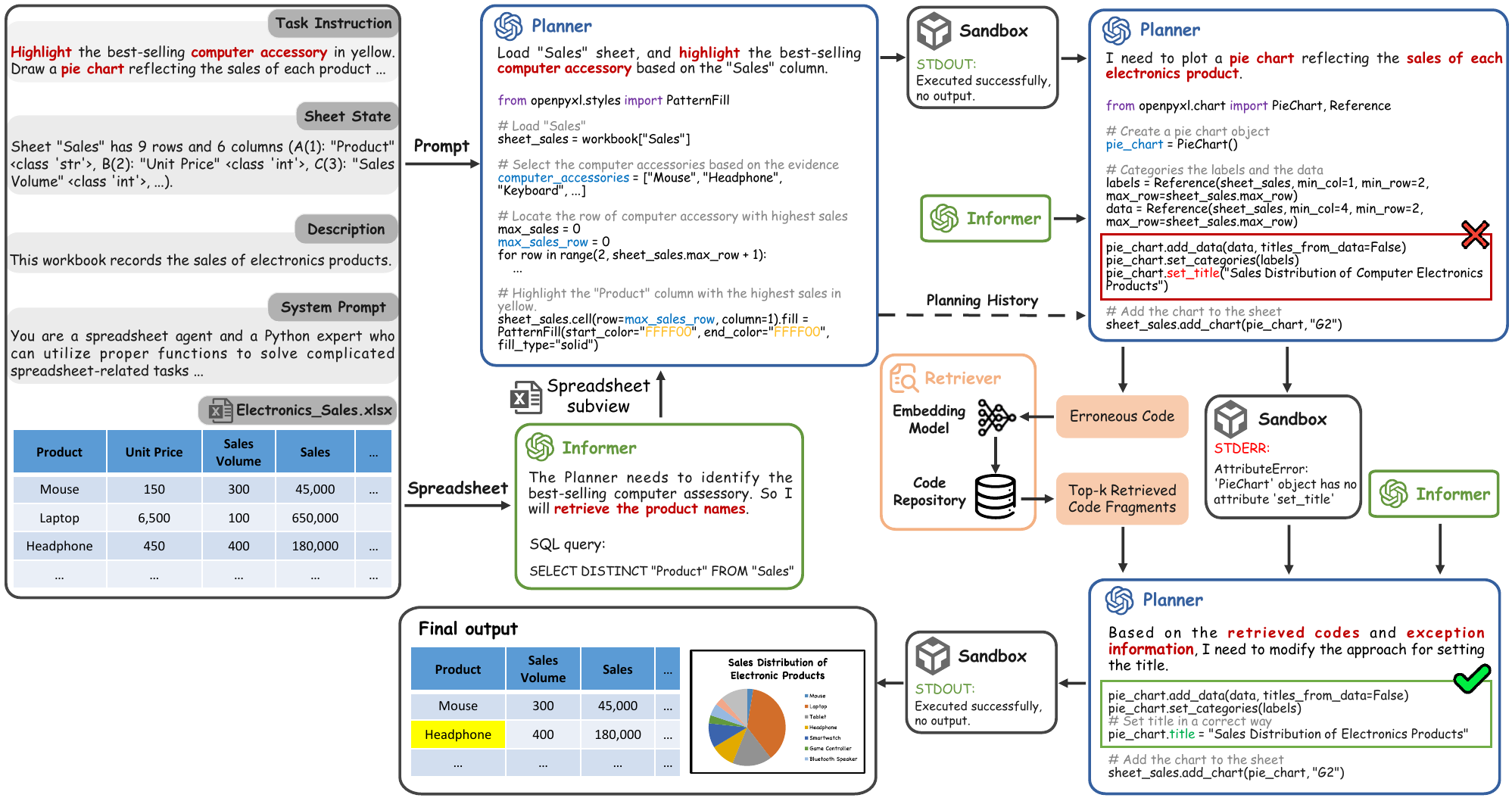}
\caption{\textbf{An illustration of SheetAgent.} SheetAgent comprises three key components, including the Planner, the Informer, and the Retriever. The Planner interacts with the target spreadsheet via a virtual sandbox. The Informer provides subtask-specific SQLs, the execution results of which serve as the evidence for the Planner to handle reasoning challenges. The Retriever is invoked to retrieve similar tutorial code snippets upon encountering an error, effectively correcting the error.}
\label{fig:framework}
\end{figure*}

\subsection{Automatic Evaluation}
\label{sec:automatic evaluation}
SheetCopilot \cite{li2024sheetcopilot} introduces a feasible method that determines task fulfillment by evaluating the alignment of key properties between the processed spreadsheet and ground truth candidates. However, this method fails to evaluate the accuracy of each operation involved, as a task may comprise multiple detailed intermediate steps. To address this challenge, we develop an automatic evaluation system that is model-free and tailored for each fine-grained operation. The advantage of that is that we are able to evaluate the performance of intermediate sub-task processes. A checklist is crafted for each task instruction. As illustrated in \cref{fig:benchmark}, within the checklist, each evaluation item comprises a \texttt{(Criterion, Sheet Index, Area, Properties)} pair. We locate the comparison region in the target spreadsheet by \texttt{(Sheet Index, Area)}. Then, corresponding \texttt{Criterion} is applied to verify whether the region aligns with the \texttt{Properties}. This design enables a detailed evaluation of LLMs' capabilities by assessing each step of task execution.

\section{SheetAgent Framework}
To quantify the challenges posed by SheetRM, we introduce an LLM-based agent framework SheetAgent. As outlined in \cref{fig:framework}, SheetAgent consists of three key components: the Planner, the Informer, and the Retriever. The Planner generates Python code to manipulate the target spreadsheet, reasoning and acting to address complex tasks. The Informer supplies task-specific SQL queries, whose execution results provide crucial subviews of the spreadsheet, narrowing the reasoning scope for the Planner and enhancing its ability to tackle complicated spreadsheet reasoning challenges. At each decision step, the Planner can perform more precise manipulations with subviews. When the Planner generates an incorrect solution, the Retriever is activated to fetch high-quality code examples from our curated repository, assisting the Planner in making more effective corrections.

\subsection{Proficient Spreadsheet Manipulation with Planner}
We design a Planner module to manipulate spreadsheets in SheetAgent. The way to interact with spreadsheets determines the precision of manipulation. Unlike SheetCopilot \cite{li2024sheetcopilot}, which uses a set of language APIs, we adopt a code-centric approach to control spreadsheets. We find Python, compared with VBA, is more suitable for manipulating spreadsheets due to its alignment with the training corpora of most existing LLMs \cite{chen2021evaluating, roziere2023code}. This Python code-centric approach reduces the occurrence of hallucinations of LLMs.

Complex spreadsheet manipulation tasks often involve multiple steps. Achieving precise control over spreadsheets is challenging without an effective feedback mechanism. To address this, we devise a closed-loop planning process where the Planner interacts with the target spreadsheet, incorporating feedback and reflection. We first concatenate task instruction $I$, system prompt $P^P$, description $D$, and the initial sheet state $s_0$ (row and column count, headers, and data type of each column) as the input. Given a snapshot of the target spreadsheet at step $t$, the Planner generates action $a_t = \text{Planner}(a_t | I, P^P, D, s_t, h_{t-1})$, where $h_{t-1}$ is the planning history. The action is evaluated in a sandbox with the feedback $o_t = \text{Sandbox}(a_t)$. If an error occurs, the Planner reflects and generates an adjusted action $a_t^* = \text{Planner}(a_t^* | I, P^P, D, s_t, h_{t-1}, o_t)$. Otherwise, the action is performed on the target spreadsheet. The spreadsheet is updated to a new state of $s_{t+1}$. The planning history is also updated to $h_t = (h_{t-1}, o_t, a_t)$. By this, Planner can achieve accurate manipulation with only the key information (i.e., the sheet state) of the spreadsheet rather than reading all the sheet data.

\subsection{Accurate Spreadsheet Perception with Informer}
Merely being aware of the sheet state does not equip the Planner to address the reasoning challenges shown in \cref{fig:benchmark}. For instance, to fulfill the instruction illustrated in \cref{fig:framework}, the Planner needs to discern which products qualify as computer accessories. However, the Planner struggles to query the spreadsheet effectively due to the absence of efficient mechanisms like SQLs and lacks the intrinsic ability to comprehend the data’s semantics. A feasible approach is constantly feeding the full spreadsheet into the Planner. However, considering the continuity of operations in a complicated task, a spreadsheet may experience multiple modifications, making it challenging to maintain a synchronized state within the Planner due to the token limit.

Therefore, we introduce the Informer to handle spreadsheets with arbitrary length and dynamic changes. Informer generates task-specific SQLs to perform queries. Initially, the tabular data in the target spreadsheet is extracted and stored in a lightweight database. At step $t$, the Informer’s objective is to select entries that align best with both the task instruction and the current step. To achieve this, we formulate the input of Informer by combining the system prompt $P^I$, the task instruction $I$, and previous actions $A_{t-1}=(a_1, \cdots, a_{t-1})$ performed by the Planner. Then, Informer generates a SQL query $q_t = \text{Informer}(q_t | I, P^I, A_{t-1}, s_t)$. $A_{t-1}$ is the reasoning trace of the Planner, enabling the Informer to generate more task-specific and robust SQLs. As shown in \cref{fig:framework}, the execution result of the SQL query is a spreadsheet subview, which serves as evidence $e_t$ for the Planner to reason over. This allows Planner to more accurately and efficiently perceive the target spreadsheet from key evidence, thereby addressing the reasoning challenges. Each time the Planner performs an operation, the spreadsheet in the database is updated to maintain synchronization.

\subsection{Robust Solution Generation with Retriever}
\label{sec:retriever}
The Retriever advises the Planner during task planning, augmenting error corrections by sourcing relevant code from a code repository. We collect high-quality code from GitHub and craft tutorial examples for each manipulation category shown in \cref{fig:benchmark}. We organize them into a compilation of code files. To improve efficiency, we employ Milvus \cite{2021milvus}, an open-source vector database, as the code repository. To construct this repository, a sliding window technique is applied to traverse these files, extracting continuous lines of code within the window size. These code fragments $\boldsymbol{\mathcal{C}}_{repo}$ are embedded into a set of vectors and stored in the code repository. The Retriever is invoked when the sandbox emits an error signal. We seek top-$k$ similar code snippets $\boldsymbol{\mathcal{C}}_{ret}^k$ as follows:
\begin{equation*}
\label{eq:retriever}
\begin{aligned}
       \boldsymbol{\mathcal{C}}_{ret}^k =\Big\{ &\mathcal{C}_{repo}^i| \mathcal{C}_{repo}^i \in \boldsymbol{\mathcal{C}}_{repo}, \forall \mathcal{C}_{repo}^j \notin \boldsymbol{\mathcal{C}}_{ret}^k, \\ & \text{sim}\left(\mathcal{E}( \mathcal{C}_q ), \mathcal{E}(\mathcal{C}_{repo}^i) \right) > \text{sim}\left(\mathcal{E}( \mathcal{C}_q ), \mathcal{E}(\mathcal{C}_{repo}^j) \right)\Big\}\text{,} 
\end{aligned}
\end{equation*}
wherein $| \boldsymbol{\mathcal{C}}_{ret}^k | = k$, $\mathcal{C}_q$ refers to the erroneous code snippet, and $\text{sim}(\cdot)$ denotes cosine similarity. The embedding function $\mathcal{E}(\cdot)$ can be represented by any pretrained language model. Consequently, the top-k similar code snippets $\boldsymbol{\mathcal{C}}_{ret}^k$ arranged in descending order are retrieved. These code snippets boost the replanning process of the Planner with $a_t^* = \text{Planner}(a_t^* | I, P^P, D, s_t, h_{t-1}, o_t, \boldsymbol{\mathcal{C}}_{ret}^k)$ for generating more robust and reliable solutions. We provide details of code collection in \cref{app:detail_retriever}.

\section{Experiment}
\label{sec:expr}
We conduct experiments on various tasks to answer the following research questions (RQs):
\begin{itemize}[leftmargin=*]
\item \textbf{Versatility (RQ1):} Is SheetAgent adept at both spreadsheet manipulation and reasoning?
\item \textbf{Universality (RQ2):} Can different LLMs benefit from the design of SheetAgent? 
\item \textbf{Difficulty (RQ3):} Why SheetRM is a challenging benchmark for existing methods?
\item \textbf{Ablation (RQ4):} How do the modules within SheetAgent contribute to its overall effectiveness?
\end{itemize}

\begin{table*}[!t]
    \centering
    \begin{minipage}{.60\textwidth}
        \centering
        \caption{Performance comparison of different methods on SCB and our SheetRM. * denotes results on a subset of SCB with 20 representative tasks. Best results are \textbf{bolded} and second-best results are \underline{underlined}.}
        \label{tab:res_scb_sheetrm}
        \begin{adjustbox}{width=\columnwidth}
        \begin{tabular}{lccccc}
        \toprule
                                                            & \multicolumn{2}{c}{\textbf{SCB}}        & \multicolumn{3}{c}{\textbf{SheetRM}}             \\ \cmidrule(lr){2-3}  \cmidrule(lr){4-6} 
        \textbf{Method} &
          \multicolumn{1}{l}{\textbf{Exec@1 \up}} &
          \multicolumn{1}{l}{\textbf{Pass@1 \up}} &
          \multicolumn{1}{l}{\textbf{Exec@1 \up}} &
          \multicolumn{1}{l}{\textbf{Pass@1 \up}} &
          \multicolumn{1}{l}{\textbf{SubPass@1 \up}} \\ \midrule
        VBA (GPT-3.5)                                       & 77.8          & 37.1          & 56.2          & 2.8           & 14.5          \\ \midrule
        OS-Copilot (GPT-4)                   & /             & \underline{60.0}*          & 74.8          & 20.5          & 56.6          \\ \midrule
        SheetCopilot (GPT-3.5)                                      & 87.3          & 44.3          & 68.1          & 0           & 16.2        \\
        SheetCopilot (GPT-4)                   & 65.0*          & 55.0*          & 52.7          & 2.2             & 30.7          \\ \midrule
        \rowcolor{lb} SheetAgent (GPT-3.5)                        & \textbf{94.1} & \textbf{61.1} & 92.4 & 31.2    & 69.8 \\
        \quad - \textit{w/o} Informer+Retriever & \underline{88.7}          & \underline{50.7}          & 88.6          & 11.4          & 57.5 \\
        \rowcolor{lb} SheetAgent (GPT-4)     & \textbf{90.0}* & \textbf{70.0}* & 89.3    & 44.8 & 77.0 \\
        \rowcolor{lb} SheetAgent (GPT-4o)     & / & / & \underline{93.1}    & \underline{46.1} & \underline{79.8} \\
        \rowcolor{lb} SheetAgent (GPT-o1)     & / & / & \textbf{96.2}    & \textbf{62.5} & \textbf{87.5} \\
        \bottomrule

        \end{tabular}
        \end{adjustbox}
    \end{minipage}%
    \hspace{.01\textwidth}
    \begin{minipage}{.38\textwidth}
        \centering
        \caption{Results of different methods on three table reasoning benchmarks. Best results are \textbf{bolded} and second-best results are \underline{underlined}.}
        \label{tab:res_reasoning}
        \begin{adjustbox}{width=0.92\columnwidth}
        \begin{tabular}{lccc}
        \toprule
        \textbf{Method}                     & \textbf{WTQ}         & \textbf{TabFact}      & \textbf{FeTaQA} \\ \midrule
        \textbf{\small{Fine-tuning based LLMs}}     & \multicolumn{1}{l}{} & \multicolumn{1}{l}{} &                \\
        \quad TAPEX \cite{liu2021tapex}           & 57.5                 & 84.2                 & 34.7           \\
        \quad UnifiedSKG \cite{xie2022unifiedskg} & 49.3                 & \textbf{85.4}        & 33.4           \\
        \quad OmniTab \cite{jiang2022omnitab}     & \underline{62.8}           & 82.8                 & \underline{34.9}     \\
        \textbf{\small{Prompting based LLMs}}       &                      &                      &                \\
        \quad GPT-3 CoT \cite{chen2023large}      & 45.7                 & 76.0                 & 27.0           \\
        \quad Binder \cite{cheng2022binding}      & 59.9                 & 82.9                 & 31.6           \\
        \quad DATER \cite{ye2023large}            & 61.6                 & 80.7                 & 30.9           \\
        \quad StructGPT \cite{jiang2023structgpt} & 52.2                 & 81.2                 & 32.5           \\ \midrule
        \rowcolor{lb} SheetAgent (GPT-3.5)                         & \textbf{64.4}        & \underline{84.8}           & \textbf{36.7}  \\ \bottomrule
        \end{tabular}
        \end{adjustbox}
    \end{minipage}
\end{table*}

\subsection{Experiment Setup}
\label{sec:expr_setup}
\textbf{Dataset and Evaluation Metrics.} We evaluate our approach SheetAgent on 5 diverse benchmarks. \textit{SheetCopilot Benchmark} (SCB), a benchmark consisting of 221 tasks, is selected to mainly assess the manipulation ability. To measure the reasoning capability, we adopt three table reasoning tasks, including \textit{WikiTableQuestions} (WTQ) \cite{pasupat2015compositional}, \textit{FeTaQA} \cite{nan2022fetaqa}, and \textit{TabFact} \cite{chen2019tabfact}. We report the performance on these tasks using their official evaluation pipeline. The 317 tasks in our \textit{SheetRM} is used to comprehensively evaluate manipulation and reasoning capabilities. Refer to \cref{app:dataset_detail} for more details of these datasets. For manipulation tasks, we adopt Exec@1 and Pass@1 following SheetCopilot. Exec@1 measures the percentage of solutions without exceptions during execution. Pass@1 is used to evaluate the successful accomplishment of the task. In addition, we use the SubPass@1 to count the success rate of subtasks in each task to assess the instruction following capability. As for reasoning tasks, we chose distinct evaluation metrics. For WTQ and TabFact, accuracy is adopted as the evaluation metric. For FeTaQA, we report the sacreBLEU score \cite{post2018call}.

\noindent\textbf{Baselines.} For SCB, we compare SheetAgent with SheetCopilot \cite{li2024sheetcopilot} and OS-Copilot \cite{wu2024copilot}, two LLM-based agent frameworks, and VBA \cite{li2024sheetcopilot}, a method that generates and runs VBA code. For table-based reasoning tasks, we select fine-tuning based LLMs like TAPEX \cite{liu2021tapex} and OmniTab \cite{jiang2022omnitab}, and prompting-based LLMs such as DATER \cite{ye2023large} and StructGPT \cite{jiang2023structgpt}. To our knowledge, there is a lack of methods capable of comprehensive spreadsheet manipulation. Besides, various approaches~\cite{dong2020learning, singh2023format5, gulwani2012spreadsheet} finetune LLMs for specific tasks like formatting and formula prediction but lack open source weights. Therefore, we mainly compare VBA, SheetCopilot, and OS-Copilot on SheetRM. We chose JSON as the table representation for its superior performance as shown in \cref{tab:table_repr}. See \cref{app:imp_detail} for implementation details.

\subsection{Versatility (RQ1)}
\label{sec:rq1}
To answer RQ1, we conduct various experiments on both spreadsheet manipulation and reasoning tasks. \cref{tab:res_scb_sheetrm} shows the results for SCB. Using GPT-3.5 as the backbone, we observe that SheetAgent outperforms SheetCopilot with a remarkable \textbf{16.8} higher Pass@1. Even without the Informer and Retriever components, our method still surpasses others in both metrics. This indicates that the generated Python code is more robust and reliable compared to VBA code or language APIs. Following \citet{li2024sheetcopilot}, we use GPT-4 on a subset of SCB, including 20 tasks. Our SheetAgent also outperforms SheetCopilot and OS-Copilot with \textbf{15.0} and \textbf{10.0} higher Pass@1 respectively. These results demonstrate that SheetAgent can better leverage the power of LLMs to achieve more accurate spreadsheet manipulation.

\begin{figure*}[!t]
\centering
\includegraphics[width=0.85\textwidth]{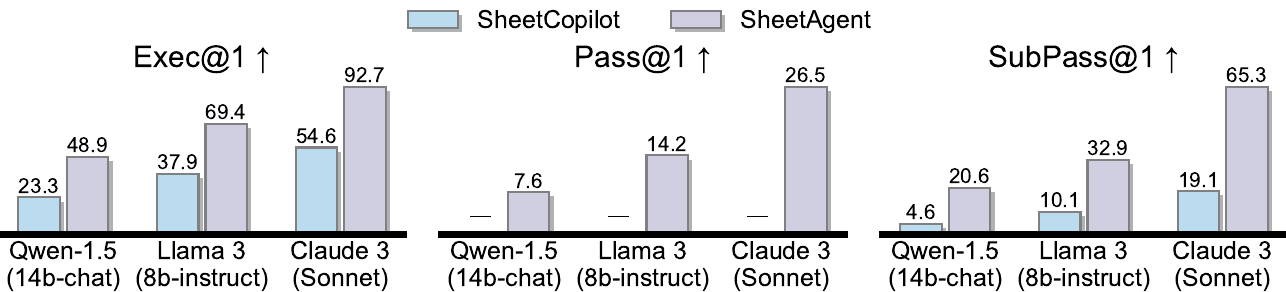}
\caption{Performance on SheetRM for other LLM backbones. ``—'' means Pass@1=0. These backbones benefit significantly from the design of SheetAgent compared to SheetCopilot.}
\label{fig:res_backbone_sheetrm}
\end{figure*}

\begin{figure*}[!t]
\centering
\subcaptionbox{}{\includegraphics[height=3.6cm,keepaspectratio]{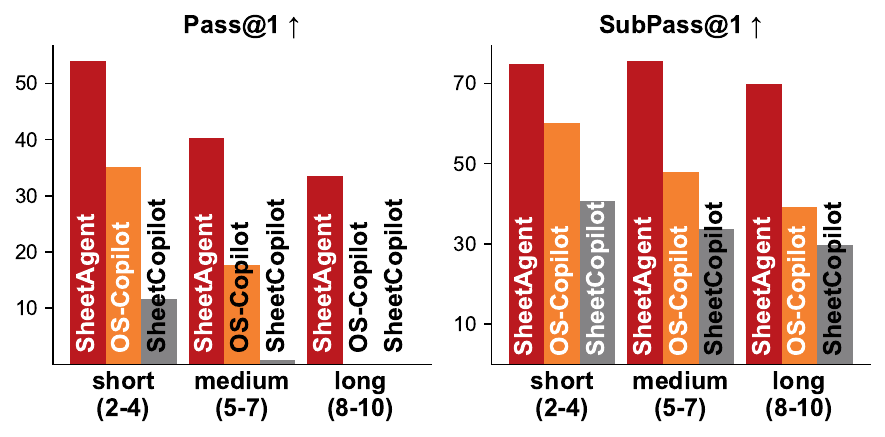}}
\subcaptionbox{}{\includegraphics[height=3.6cm,keepaspectratio]{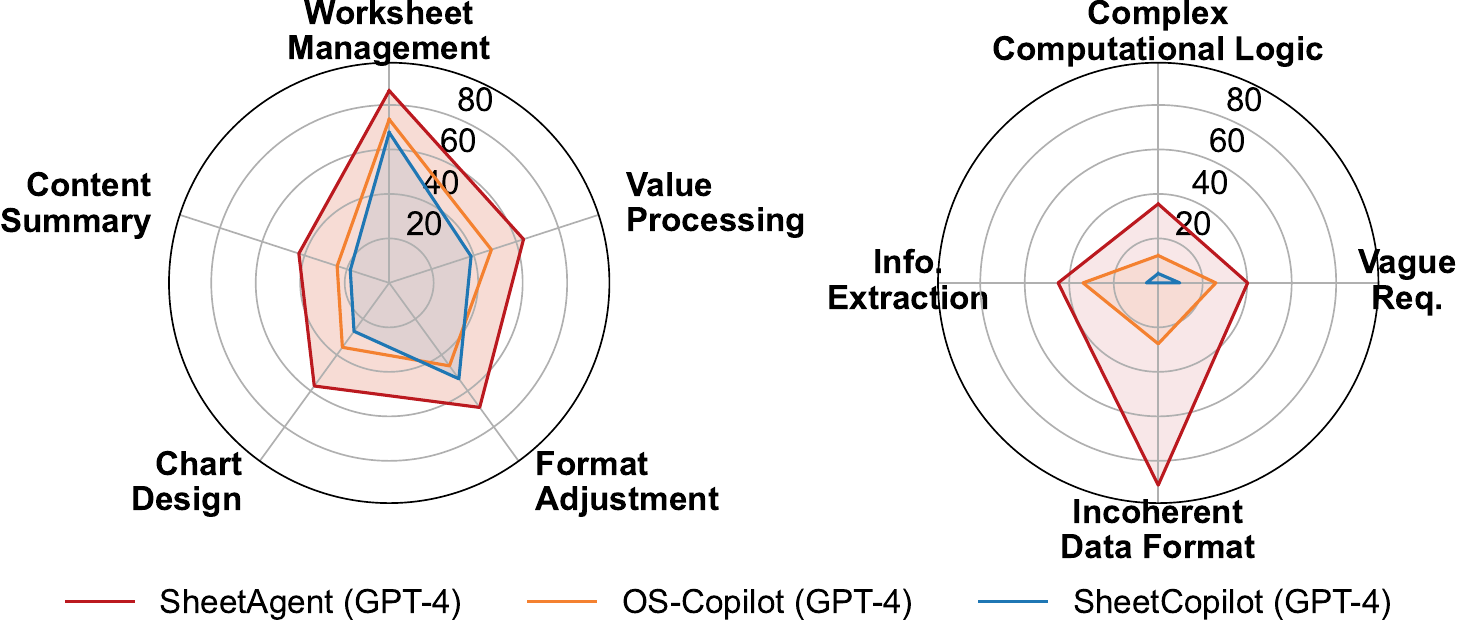}}
\caption{Comparison between SheetAgent and SheetCopilot with GPT-4. (a) Comparison of Pass@1 and SubPass@1 under different task horizon levels. (b) Pass rate of different manipulation categories (left) and reasoning challenges (right).}
\label{fig:task_len_cat_acc}
\end{figure*}

Further experiments focus on assessing SheetAgent's reasoning capability. We remove the Retriever as these tasks typically involve simpler operations like sorting and filtering. Results in \cref{tab:res_reasoning} show that SheetAgent outperforms other baselines on WTQ and FeTaQA tasks, indicating its capability to provide precise and informative responses.  Besides, SheetAgent surpasses all fine-tuning based methods, and achieves comparable performance as SOTA method UnifiedSKG \cite{xie2022unifiedskg} on TabFact. The results underscores the synergy between the Planner and Informer, which significantly enhances SheetAgent's efficacy in table reasoning tasks. Full results are provided in \cref{app:res_reasoning_full}.

We compare SheetAgent with baselines on the SheetRM to evaluate reasoning and manipulation capabilities.
For fair comparison, we have improved SheetCopilot based on the open-source version with error feedback functionality. Results in \cref{tab:res_scb_sheetrm} show SheetAgent significantly outperforms other baselines in three aspects: (1) \textbf{Robust solution generation:} SheetAgent achieves an Exec@1 of \textbf{92.4} with GPT-3.5, indicating more robust solutions, meaning Python code generated by LLMs is more robust than VBA and language APIs. (2) \textbf{Strong manipulation proficiency:} SheetAgent is proficient in complex multi-category tasks, achieving a maximum SubPass@1 of \textbf{77.0}, more than double that of SheetCopilot with GPT-4. (3) \textbf{Advanced reasoning ability:} SheetAgent can solve more reasoning challenges, whereas SheetCopilot struggles significantly (Pass@1 \textbf{44.8 vs 2.2}). This reflects the superior reasoning capabilities of SheetAgent.
Additionally, we evaluate our SheetAgent using the latest GPT-4o and GPT-o1. As the model capabilities increase, it is observed that SheetAgent achieves better performance on the SheetRM benchmark, with Pass@1 scores of \textbf{46.1} and \textbf{62.5}, respectively.
We also provide an illustrative case in \cref{app:case_compare} to further demonstrate why SheetAgent outperforms SheetCopilot in tasks with reasoning challenges.

\subsection{Universality (RQ2)}
\label{sec:rq2}
To answer RQ2, we compare our SheetAgent with SheetCopilot across various LLM backbones on SheetRM. As presented in \cref{tab:res_scb_sheetrm} and \cref{fig:res_backbone_sheetrm}, SheetAgent shows remarkable improvements in all evaluated metrics on diverse backbones such as GPTs and Claude. Despite with smaller, open-source backbones, we can continue to observe the same results. Furthermore, SheetCopilot fails to pass any task completely on open-source models, and possesses lower Exec@1 scores, highlighting its challenges in generating feasible solutions. These results confirm the universality of SheetAgent, illustrating its consistent performance improvements across different LLM backbones regardless of scale. Meanwhile, we note that the differences between various LLM backbones mainly stem from their fundamental capabilities, such as instruction following and code generation abilities.
We further analyze failure cases on SheetRM in \cref{app:failure_case} to identify the deficiencies and opportunities for improvement in SheetAgent under different LLM backbones.

\subsection{Difficulty (RQ3)}
We explore the challenges of our proposed SheetRM benchmark from three perspectives: task horizon, task categories, and reasoning challenges. We compare SheetAgent with SheetCopilot and OS-Copilot against the same GPT-4 backbone. Tasks are categorized into three levels based on their horizon: short (2-4), medium (5-7), and long (8-10). As depicted in \cref{fig:task_len_cat_acc}(a), both methods exhibit a decreasing trend in Pass@1 and SubPass@1 as task horizon increases, indicating the difficulty of long-horizon tasks in our benchmark. Furthermore, \cref{fig:task_len_cat_acc}(b) presents the performance across different manipulation categories and reasoning challenges by evaluating subtask success rates. Both methods struggle with more complex tasks like chart design and content summary. Additionally, SheetCopilot can hardly address reasoning challenges.
These findings highlight the challenges SheetRM poses in domains requiring consistent, robust reasoning and manipulation.

\subsection{Ablation (RQ4)}
\label{sec:ablation}
\textbf{Effects of Each Module.} \cref{tab:component_abalation} reveals the effects of SheetAgent modules. Pass@1 drops dramatically without Informer, indicating its vital role in handling reasoning challenges by providing relevant information. Exec@1 also decreases sharply without Retriever, showing that high-quality examples help the Planner generate reliable code. Without both Informer and Retriever, SheetAgent performs poorest, highlighting the need for both reasoning and manipulation capabilities to tackle complex tasks effectively. Combining the results in \cref{tab:res_scb_sheetrm}, even with only the Planner, SheetAgent performs decently compared to SheetCopilot, showcasing the benefits of a code-centric approach.

\noindent\textbf{Table Representations.} Tabular data requires reliable representations for LLMs to recognize attribute relationships. We ablate four table representations—JSON, DFLoader, Markdown, and HTML—for SheetAgent on WTQ and SheetRM. Results in \cref{tab:table_repr} show JSON outperforms other formats. HTML performs poorly on SheetRM due to verbosity and token limits. We provide illustration of different representations in \cref{app:table_repr} with in-depth analysis. More additional ablations about temperature and vision-enabled SheetAgent can be found in \cref{app:additional_res}.

\begin{table}[!t]
\centering
\caption{Ablation study of different proposed components in SheetAgent on SheetRM dataset.}
\label{tab:component_abalation}
\resizebox{\columnwidth}{!}{%
\begin{tabular}{lccc}
\toprule
\textbf{Method}             & \textbf{Exec@1 \up} & \textbf{Pass@1 \up} & \textbf{SubPass@1 \up} \\ \midrule
\rowcolor{lb} SheetAgent (GPT-3.5)                    & 92.4               & \textbf{31.2}               & \textbf{69.8}                  \\
\quad -\textit{w/o} Informer           & \textbf{95.3} \textcolor{dg}{(+2.9)}             & 13.2 \textcolor{red}{(-18.0)}              & 65.8 \textcolor{red}{(-4.0)}                  \\
\quad -\textit{w/o} Retriever          & 84.2 \textcolor{red}{(-8.2)}            & 19.9 \textcolor{red}{(-11.3)}               & 63.7 \textcolor{red}{(-6.1)}                  \\
\quad -\textit{w/o} Informer+Retriever & 88.6 \textcolor{red}{(-3.8)}             & 11.4 \textcolor{red}{(-19.8)}               & 57.5 \textcolor{red}{(-12.3)}                  \\ \bottomrule
\end{tabular}%
}
\end{table}

\begin{table}[!t]
\centering
\caption{Ablation study on different representations. Best results are \textbf{bolded} and suboptimal results are \underline{underlined}.}
\label{tab:table_repr}
\resizebox{\columnwidth}{!}{%
\begin{tabular}{lcccc}
\toprule
\multirow{2}{*}{\textbf{Representation}} & \multirow{2}{*}{\textbf{WTQ}} & \multicolumn{3}{c}{\textbf{SheetRM}}    \\ \cmidrule(lr){3-5} 
                                &                      & \textbf{Exec@1 \up} & \textbf{Pass@1 \up} & \textbf{SubPass@1 \up} \\ \midrule 
JSON                            & \textbf{63.3}                 & \textbf{92.4}      & \textbf{31.2}      & \textbf{69.8}         \\
DFLoader                        & 59.7                 & \underline{90.9}     & \underline{30.0}      & \underline{67.5}         \\
Markdown                        & 58.6                & 90.2      & 29.0      & 65.4         \\
HTML                            & \underline{62.1}                 & 85.2      & 23.3      & 57.5         \\ \bottomrule
\end{tabular}%
}
\end{table}

\section{Related Work}
\textbf{LLMs for Table Reasoning.}
Recent research \cite{wei2022chain, wangself} has demonstrated the excellent ability of LLMs for table reasoning tasks. \citet{chen2023large} showcased that LLMs like GPT-3 \cite{brown2020language} are capable of reasoning over tables. Binder \cite{cheng2022binding} leverages Codex \cite{chen2021evaluating} to generate executable SQL programs to answer table-based questions. DATER \cite{ye2023large} decomposes the table and question into finer granularity descriptions through Codex. StructGPT \cite{jiang2023structgpt} designs an LLM-based framework for structured data and uses it for table question answering. However, these methods are tailored for tasks like question answering or fact verification, typically involving direct queries or explicit statements. As a result, they struggle to handle long-horizon manipulation tasks because of dynamic changes and token limits.

\noindent\textbf{Automatic Spreadsheet Manipulation.}
Early research \cite{gulwani2011automating, gulwani2012spreadsheet, gulwani2014nlyze, balog2016deepcoder} focus on leveraging program synthesis to guide spreadsheet manipulation. However, these methods fail to generate effective programs without high-quality query specifications. To address this, some work \cite{dong2020learning, chen2021spreadsheetcoder, singh2023format5, he2023hermes} employ deep learning methods to automate spreadsheet manipulation tasks. Despite excellent performance in narrow domains like formatting and formula prediction, they cannot handle a broader range of operations. Given the remarkable performance of LLMs on various tasks \cite{sun2024harnessing,zhang2024agentcf,wang2024explainable}, their use for comprehensive spreadsheet manipulation has been explored \cite{payan2023instructexcel, li2024sheetcopilot, zha2023tablegpt, zhang2024data, wu2024copilot}. \citet{payan2023instructexcel} utilizes LLMs to generate OfficeScripts code with multiple domains. SheetCopilot \cite{li2024sheetcopilot} builds an autonomous agent for invoking custom APIs to manipulate spreadsheets.
TableGPT \cite{zha2023tablegpt} fine-tunes an LLM to understand and operate on tables using external functions. OS-Copilot \cite{wu2024copilot} proposes an OS-oriented agent framework that automates spreadsheet manipulation. However, both approaches simplify real-world requirements and overlook key reasoning challenges such as unclear expressions and multi-step logic. Unlike them, we further explore these real-life reasoning challenges and propose a collaborative agent framework that explicitly tackles spreadsheet tasks requiring deeper reasoning.

\section{Conclusion}
In this work, we introduce SheetRM, a more complex and realistic benchmark designed to evaluate the capabilities of agents in performing precise spreadsheet manipulations that require advanced reasoning abilities. Furthermore, We introduce SheetAgent that leverages the power of LLMs to tackle these challenging tasks. Comprehensive experiments have been conducted to assess the reasoning and manipulation proficiency of SheetAgent. We anticipate that SheetRM will serve as a cornerstone for advancing the development of sophisticated generalist agents dedicated to spreadsheet tasks. Furthermore, we hope SheetAgent can alleviate the burden of tedious sheet transactions through automated workflows. While SheetAgent demonstrates strong performance, we acknowledge several limitations like library coverage and token usage, detailed in \cref{app:limits}, which we leave as future work. We also hope to push the boundaries of automatic spreadsheet processing.

\begin{acks}
This work is supported by the National Key Research and Development Program of China (Grant No. 2024YFE0210900), the National Natural Science Foundation of China (Grant Nos. 62106172, 62422605, 92370132), the National Key Research and Development Program of China (Grant No. 2022ZD0116402) and the Xiaomi Young Talents Program of Xiaomi Foundation.
\end{acks}

\bibliographystyle{ACM-Reference-Format}
\bibliography{ref}

\appendix

\section{Details of SheetRM Benchmark}
\label{app:sheetrm_detail}
\subsection{Details of Dataset Collection}
\label{app:collection}
\textbf{Spreadsheet Collection.} The spreadsheets curated in SheetRM dare derived from an online examination question bank. We filter out files that are protected, corrupted, or otherwise inaccessible. Within each spreadsheet, the first row of each column must include a header, with the actual data entries starting from the second row. Besides, we ensure all data in each sheet begin from cell A1. We assume that these spreadsheets have already undergone a process to remove some personal information. However, to minimize privacy risk by leaking important personal information, we further implement measures to ensure that no privacy issues arise. Specifically, we modify potentially sensitive information, such as adding noise to the age data and anonymizing bookstore names to general labels like Bookstore A, Bookstore B, etc.

\noindent\textbf{Task Verification.} As mentioned in \cref{sec:data_construction}, we instruct GPT-4 to generate realistic tasks that mimic user requests adhering to four guidelines: the tasks should only involve predefined operations, cover diverse manipulation categories, exhibit a long-horizon nature by encompassing multiple subtasks, and incorporate at least one subtask that presents the specified reasoning challenges. This procedure yields a collection of 2316 subtasks. We use GPT-3.5 to filter task instructions that have a lot of semantic duplication to maintain uniqueness. After this, 1973 subtasks are reserved. Furthermore, our internal annotators verify these subtasks manually to ensure quality, which increases the probability that they will be completed by LLMs. Specifically, we adopt two strategies: (1) programming and (2) specialized software. For programming, we ask our internal annotators to write code to complete specific subtask. For specialized software, we use Microsoft Excel to solve the subtask. We accept the subtask only if both strategies solve the subtask. This cross-way validation approach guarantees the reliability of the subtasks. We obtain 1625 subtasks after this process. Finally, we combine these subtasks for different spreadsheets considering horizon and complexity, which leads to 317 task instructions.

\subsection{Comparison Between SheetRM and SCB}
\label{app:comp_sheetrm_scb}
We conclude the differences that highlight the advantages of our proposed SheetRM dataset compared with SCB as follows:
\begin{itemize}[leftmargin=*, topsep=0pt, itemsep=0pt, , parsep=0pt]
    \item \textbf{More sheets:} The number of spreadsheet files in SheetRM and SCB is comparable. Besides, SheetRM maintains more spreadsheet files and sheets than SCB \textbf{(41 vs 28 \& 137 vs 31)}. Each spreadsheet file contains more complex logical relationships and information.
    \item \textbf{More subtasks and longer task horizon:} As shown in \cref{tab:comp_sheetrm_scb}, SheetRM maintains more subtasks \textbf{(1625 vs 431)} with longer horizon tasks \textbf{(averaging 5.13 vs 1.95)}. Detailed task length distribution is presented in \cref{fig:cat_subtask}.
    \item \textbf{Broader categories and more reasonable division:} SCB categorizes tasks into 6 main types: Entry \& Manipulation, Formatting, Pivot Tables, Charts, Formulas, and Management, which results in unbalanced coverage and vague definitions. For example, Formula is basically a type of numerical computation and overlaps with Management and Manipulation, etc. In contrast, SheetRM divides the \textbf{5 major categories and 36 sub-categories} from coarse to fine and minimizes the overlap of sub-operations. We believe this allows for a better evaluation of the agents.
    \item \textbf{Finer-grained and more flexible evaluation:} We propose an \textbf{automated checklist-based evaluation} in SheetRM that is flexible and accurate for each subtask in the middle of a process, whereas SCB directly compares the final spreadsheets with the ground truth spreadsheets, ignoring the intermediate process of evaluation.
    \item \textbf{Introduction of reasoning challenges:} It is worth noting that reasoning challenges are innovatively introduced combined with manipulation in SheetRM. In real-world spreadsheet tasks, it is often necessary to reason and analyze problems and data in order to carry out operations. The SCB simplifies the task objectives by assessing only the LLM's ability to manipulate spreadsheets. Instead, our proposed SheetRM presents more realistic and challenging tasks. Please refer to \cref{app:challenges} for further elaboration.
\end{itemize}

\begin{table*}[htbp]
\centering
\caption{Comparison of statistical data between SheetRM and SCB.}
\label{tab:comp_sheetrm_scb}
\resizebox{\textwidth}{!}{%
\begin{tabular}{lccccccc}
\toprule
\textbf{Dataset Name} &
  \textbf{\# Files} &
  \textbf{\# Sheets} &
  \textbf{\# Task Instructions} &
  \textbf{\# Subtasks} &
  \textbf{Avg. of Task Length} &
  \textbf{Median of Task Length} &
  \textbf{Max Task Length} \\ \midrule
SheetRM (Ours) &
  \textbf{41} &
  \textbf{137} &
  \textbf{317} &
  \textbf{1625} &
  \textbf{5.13} &
  \textbf{5} &
  \textbf{10} \\
SCB &
  28 &
  31 &
  221 &
  431 &
  1.95 &
  2 &
  7 \\ \bottomrule
\end{tabular}
}
\end{table*}

\subsection{Detailed Statistics of Dataset}
\label{app:stat}

\begin{figure*}[htbp]
\centering
{\includegraphics[height=6cm,keepaspectratio]{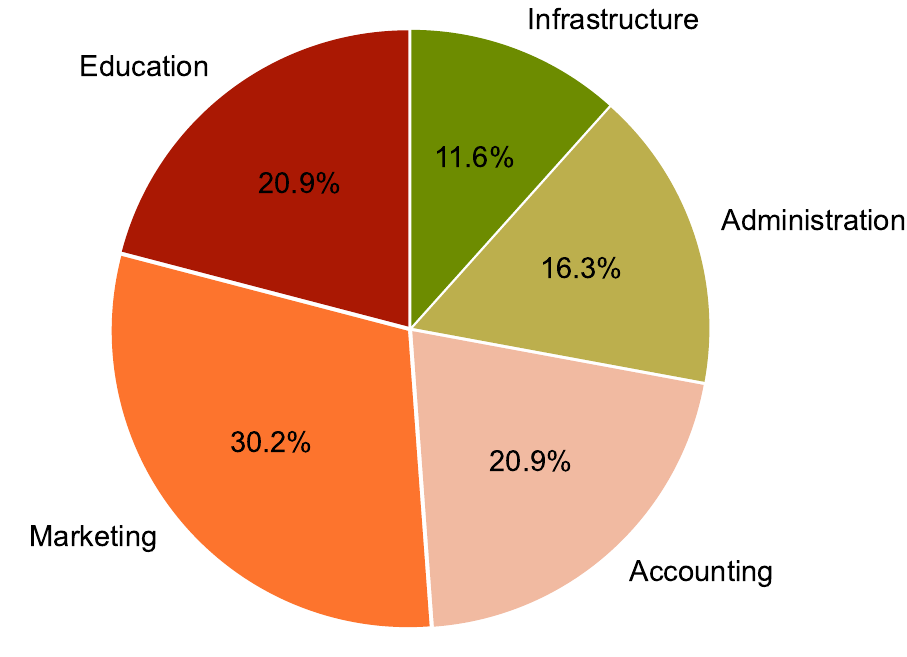}}
{\includegraphics[height=6cm,keepaspectratio]{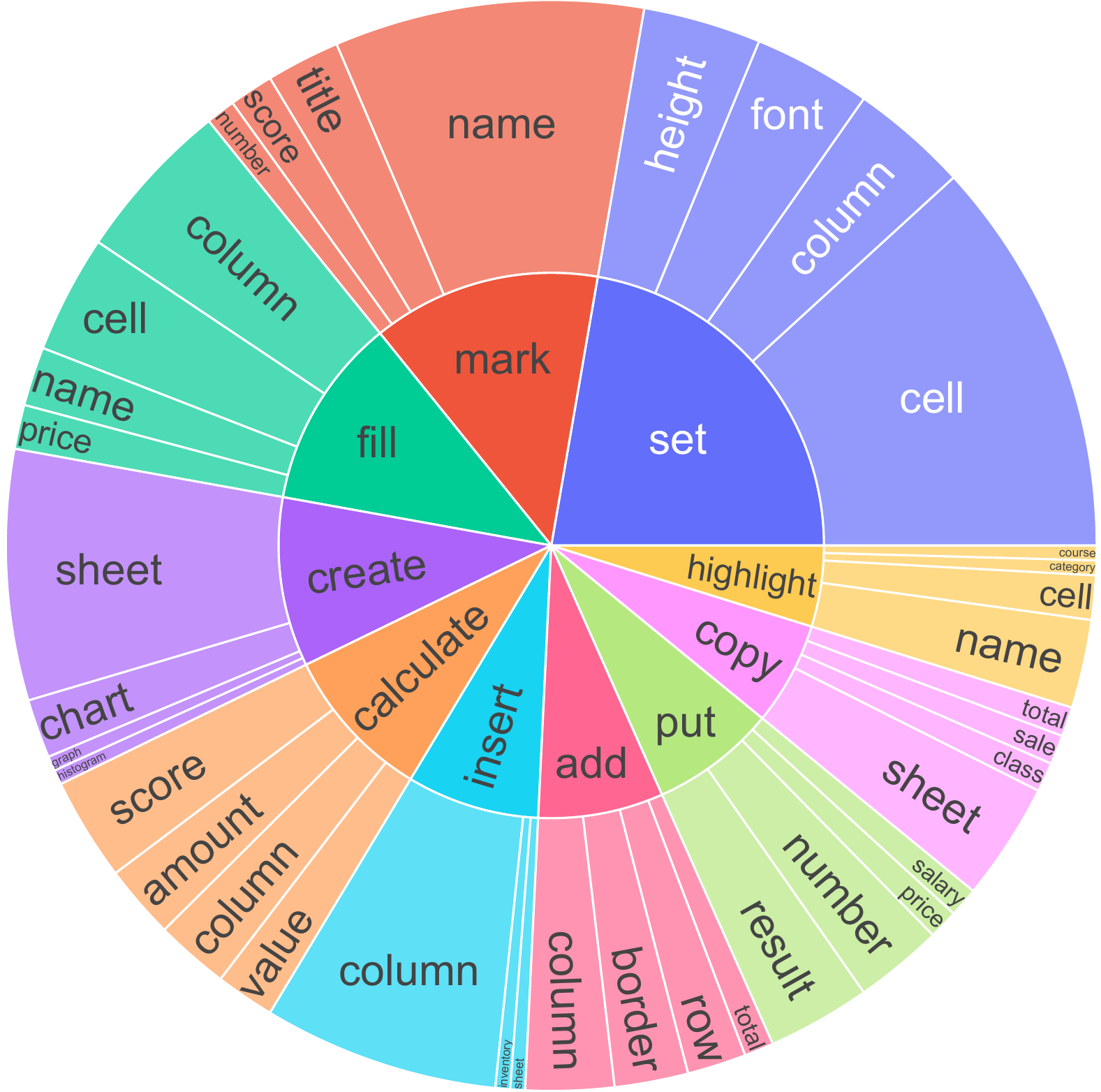}}
\caption{(Left) Distribution of fields to which the spreadsheet files belong. (Right) An illustration of verb-noun phrases in the task instructions. We count the top 10 most frequent root verbs and their associated nouns, ranking the top four for each. These verb-noun combinations showcases the diversity of the generated instructions.}
\label{fig:domain_verb}
\end{figure*}

\textbf{Spreadsheet Files.} We provide more detailed statistics of our SheetRM dataset. We collect spreadsheets covering multiple fields. As illustrated in \cref{fig:domain_verb} (Left), we categorize these spreadsheet files into five main fields, reflecting the significant areas where spreadsheets are frequently employed to handle a variety of tasks. We manually annotate a short natural language description as a summary for each spreadsheet file, aiming to stimulate inherent knowledge of LLMs. Each description provides an overview for LLMs to better understand the background information. We provide the descriptions in \cref{tab:sheet_desc}.

\begin{figure*}[htbp]
\centering
{\includegraphics[height=5cm,keepaspectratio]{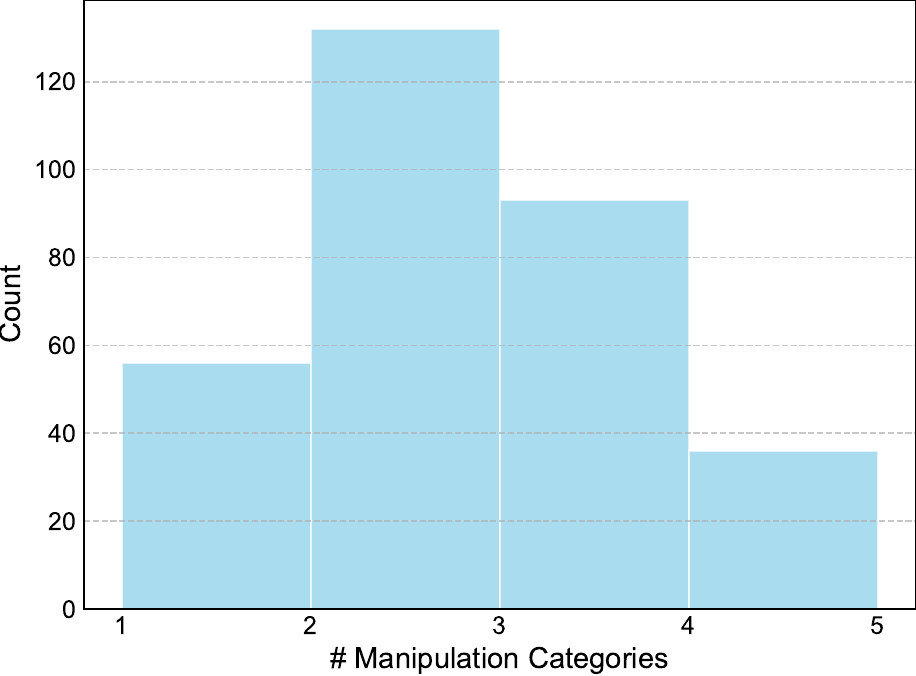}}
{\includegraphics[height=5cm,keepaspectratio]{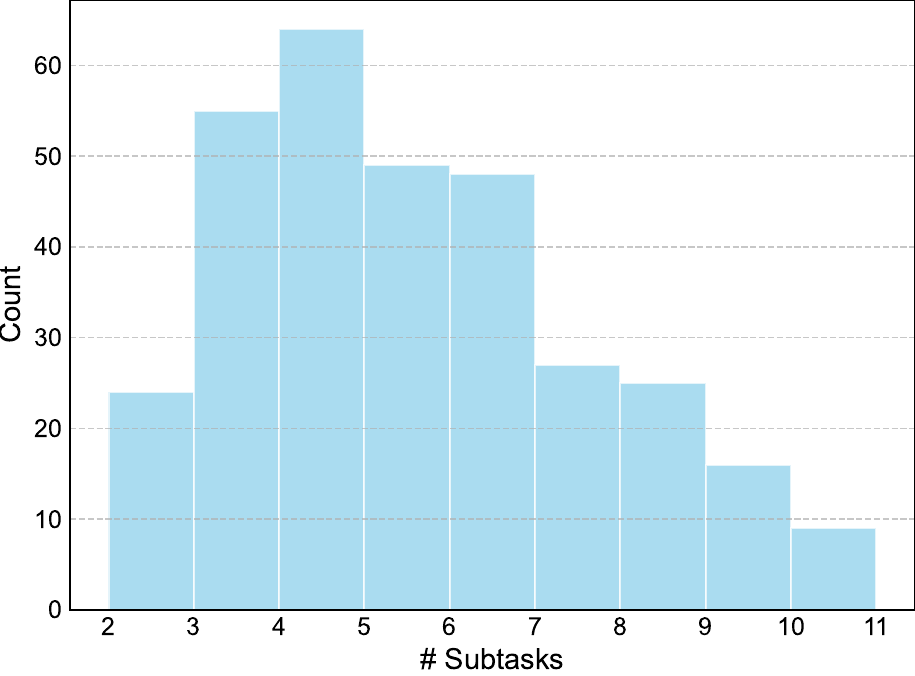}}
\caption{(Left) Distribution of manipulation categories in each task. (Right) Histogram of the task length.}
\label{fig:cat_subtask}
\end{figure*}

\noindent\textbf{Task Instruction.} We cluster the commonly used operation when working with spreadsheets into five categories, namely \textbf{Value Processing}, \textbf{Worksheet Management}, \textbf{Format Adjustment}, \textbf{Chart Design}, and \textbf{Content Summary}. For each manipulation category, we further break it down into fine-grained operations. We believe these operations can cover most spreadsheet affairs. The description of these operations is introduced in \cref{tab:oper_desc}. \cref{fig:domain_verb} (Right) demonstrates the distribution of verb-noun phrases within our 317 task instructions. We highlight the ten most frequent root verbs and their four primary associated nouns, showcasing the diversity of task instructions in the SheetRM dataset. Additionally, we show the distributions of the number of manipulation categories and subtasks for these task instructions (see \cref{fig:cat_subtask} (Left)). The majority of tasks span 2 or 3 manipulation categories, with a decent portion encompassing 4 categories, underscoring the diversity of tasks in the SheetRM dataset. We further count the number of subtasks in each task. As displayed in \cref{fig:cat_subtask}, each task includes at least 2 sub-tasks, with the most complex extending to 10. Predominantly, the tasks vary in length from 3 to 7. This reflects the long horizon feature of SheetRM, which poses a significant challenge to LLMs.

\subsection{Explanation of Reasoning Challenges}
\label{app:challenges}
Our SheetRM dataset stands out from other spreadsheet manipulation collections due to its emphasis on reasoning-dependent manipulation. Specifically, each task incorporates reasoning challenges. We draw inspiration from several popular table reasoning tasks, including table question answering datasets WikiTableQuestions and FeTaQA, and table fact verification task TabFact. We analyze cases within these datasets that most models struggled with and identify four types of reasoning challenges, namely \textbf{Complex Computational Logic}, \textbf{Vague Requirements}, \textbf{Incoherent Data Format}, and \textbf{Information Extraction}. We find that these reasoning challenges are prevalent in real-world spreadsheet manipulation tasks due to the diversity of human expression. Thus, integrating practical insights, we incorporate these reasoning challenges into our spreadsheet manipulation tasks. We elaborate these challenges with descriptions and specific examples:

\begin{tcolorbox}[title={Complex Computational Logic},breakable,enhanced,colframe=red!50!black,colback=red!10!white,
arc=1mm,colbacktitle=red!10!white,
fonttitle=\bfseries,coltitle=red!50!black,
attach boxed title to top text left=
{yshift=-0.50mm},
boxed title style={skin=enhancedfirst jigsaw,
size=small,arc=1mm,bottom=-1mm,
interior style={fill=none,
top color=red!30!white,
bottom color=red!20!white}}]
\textbf{Description:} \par
Problems that require more than one reasoning steps to be solved. \par
\textbf{Example Sheet:} \par
\begin{table}[H]
\resizebox{\columnwidth}{!}{%
\begin{tabular}{cccc}
\textbf{Name} & \textbf{Date of Entry} & \textbf{Educational Qualification} & \textbf{Salary} \\
Alice & 3/1/2001 & Master & 11,100 \\
Bob & 12/1/2006 & Bachelor & 10,350 \\
... & ... & ... & ... \\
John & 1/9/2011 & Doctor & 41,100 \\
\end{tabular}}
\end{table}
\textbf{Instruction:} \par
Which period, 2001-2006 or 2007-2012, had a higher proportion of employees with bachelor's degrees? For the period with the higher proportion, calculate the average salary of the undergraduate employees and put it in cell E1. \par
\textbf{Challenge:} \par
To fulfill this instruction, the capability of multi-step reasoning is required.
\end{tcolorbox}

\begin{tcolorbox}[title={Vague Requirements},breakable,colframe=red!50!black,colback=red!10!white,
arc=1mm,colbacktitle=red!10!white,enhanced,
fonttitle=\bfseries,coltitle=red!50!black,
attach boxed title to top text left=
{yshift=-0.50mm},
boxed title style={skin=enhancedfirst jigsaw,
size=small,arc=1mm,bottom=-1mm,
interior style={fill=none,
top color=red!30!white,
bottom color=red!20!white}}]
\textbf{Description:} \par
Problems that refer to incomplete or ambiguous specifications which lack clarity and precision, making it challenging to understand and fulfill the intended goals or objectives. \par
\textbf{Example Sheet:} \par
\begin{table}[H]
\begin{tabular}{ccc}
\textbf{BookID} & \textbf{Book Name} & \textbf{Unit Price} \\
BK-83024 & VB Programming & 38 \\
BK-83026 & Access Programming & 35 \\
... & ... & ...  \\
BK-83029 & Network Technology & 43 \\
\end{tabular}
\end{table}
\textbf{Instruction:} \par
Highlight database-related books in yellow. \par
\textbf{Challenge:} \par
To fulfill this instruction, Reasoning over the sheet contents to identify which books are relevant to the database.
\end{tcolorbox}

\begin{tcolorbox}[title={Incoherent Data Format},enhanced,colframe=red!50!black,colback=red!10!white,
arc=1mm,colbacktitle=red!10!white,
fonttitle=\bfseries,coltitle=red!50!black,
attach boxed title to top text left=
{yshift=-0.50mm},
boxed title style={skin=enhancedfirst jigsaw,
size=small,arc=1mm,bottom=-1mm,
interior style={fill=none,
top color=red!30!white,
bottom color=red!20!white}}]
\textbf{Description:} \par
Problems that arise when the description provided pertains to the spreadsheet data, yet the units or formats mentioned do not align with those represented in the spreadsheet. \par
\textbf{Example Sheet:} \par
\begin{table}[H]
\begin{tabular}{cc}
\textbf{Name} & \textbf{Date of Birth} \\
Alice & 12/27/1964  \\ 
Bob & 9/28/1974  \\
... & ...  \\
John & 7/19/1987 \\
\end{tabular}
\end{table}
\textbf{Instruction:} \par
Mark the names of employees born after 1985-1-1 in red. \par
\textbf{Challenge:} \par
To fulfill this instruction, the ``Date of Birth'' column should be inferred to align the format.
\end{tcolorbox}

\begin{tcolorbox}[title={Information Extraction},breakable,enhanced,colframe=red!50!black,colback=red!10!white,
arc=1mm,colbacktitle=red!10!white,
fonttitle=\bfseries,coltitle=red!50!black,
attach boxed title to top text left=
{yshift=-0.50mm},
boxed title style={skin=enhancedfirst jigsaw,
size=small,arc=1mm,bottom=-1mm,
interior style={fill=none,
top color=red!30!white,
bottom color=red!20!white}}]
\textbf{Description:} \par
Problems that require specific information to be extracted from the spreadsheet. \par
\textbf{Example Sheet:} \par
\begin{table}[H]
\resizebox{\columnwidth}{!}{%
\begin{tabular}{lcc}
\textbf{Venue} & \textbf{Opponent} & \textbf{Final Score} \\
Memphis, Tennessee, USA & Jim Courier & 7-5, 6-7, 6-7 \\
Australian Open, Melbourne, Australia & Pete Sampras & 6-7, 4-6, 4-6 \\
... & ... & ...  \\
Estoril, Portugal & Albert Costa & 6-2, 3-6 \\
\end{tabular}}
\end{table}
\textbf{Instruction:} \par
Extract the scores from the first round of the finals into the new column "First Round Score". \par
\textbf{Challenge:} \par
To fulfill this instruction, Information about the ``Final Score'' is required to determine how to extract the first round score.
\end{tcolorbox}

\begin{table*}[htbp]
\centering
\caption{A short natural language description of the spreadsheet files we collect in SheetRM dataset.}
\label{tab:sheet_desc}
\resizebox{\textwidth}{!}{%
\begin{tabular}{lp{0.75\textwidth}}
\toprule
\textbf{Spreadsheet File} & \textbf{Description}                                                                                                                    \\ \midrule
BookSales                 & This workbook presents data related to book sales.                                                                                      \\
StudentsGrade             & This workbook is about organizing and analyzing student transcripts for first-grade students.                                           \\
ABProductSales            & This workbook presents data related to product A and B.                                                                                 \\
Reimbursement             & The workbook shows the company's travel expense reimbursement status for the year 2013.                                                 \\
ElectronicsSales          & The workbook is about conducting statistical analysis of the company's sales.                                                           \\
PayrollSummary            & The workbook is the March 2014 employee salary sheet.                                                                                   \\
TeachingFees              & This workbook shows the teaching situation and instructor hourly fees for the courses in the Teaching Research Office in the year 2012. \\
Deposit                   & The workbook is a bank deposit journal.                                                                                                 \\
ComputerBookSales         & The workbook depicts the sales figures for computer-related books in December 2012.                                                     \\
ScienceMajorGrade         & The workbook shows the final exam grades for the Information and Science major.                                                         \\
PersonnalInformation      & This workbook is the personnel file information of company employees.                                                                   \\
ComputerBookSales2        & This workbook represents the sales statistics of computer-related books.                                                                \\
AppliancesSales           & This workbook shows the sales statistics of various household appliances.                                                               \\
DepartmentSales           & This workbook documents the sales performance of company's products in the first half of the year.                                      \\
QuartersSales             & This workbook summarizes the sales performance for the first two quarters.                                                              \\
FinalGrade                & This workbook provides a detailed analysis of students' final grades.                                                                   \\
ParkingFees               & This workbook keeps track of parking fees and the associated rates.                                                                     \\
LivingCosts               & This workbook displays an individual's monthly expense report.                                                                          \\
StudentsGrade2            & This workbook displays the grades for each subject in the class.                                                                        \\
LawMajorGrade             & This workbook presents the final grade analysis of law students from the 2012 cohort.                                                   \\
YearsSales                & This workbook documents the sales statistics of company products in 2012 and 2013.                                                      \\
YearEndSalary            & This workbook provides the year-end salary details of employed staff members.                                                           \\
AirQuality                & This workbook illustrates the air quality data for major cities in China.                                                               \\
SalesAndPurchase          & This workbook is a record of this year's sales and purchase data.                                                                       \\
PersonnelChange           & This workbook contains the personal details of company employees for the year 2019, including their entry and departure information.    
                          \\
ProductLaunchPlan    & This workbook outlines the   product launch timeline, key milestones, and marketing strategies.                                         \\
StudentAttendance    & This workbook tracks the attendance records of students across   various grades.                                                        \\
QuarterlyEarnings    & This workbook presents the company's earnings and financial   reports for each quarter of 2020.                                         \\
OfficeInventory      & This workbook lists office supplies, including current stock   levels and reorder statuses.                                             \\
RoadMaintenanceLog   & This workbook logs the maintenance schedule and costs   associated with road repairs in the city.                                       \\
CustomerSurvey       & This workbook compiles customer feedback from recent marketing   campaigns and product surveys.                                         \\
TeacherPerformance   & This workbook evaluates teacher performance based on student   feedback and exam results.                                               \\
BudgetForecast       & This workbook forecasts the company's budget allocations for   the next fiscal year.                                                    \\
HRLeaveTracker       & This workbook tracks employee leave, including vacation days   and sick leave balances.                                                 \\
BridgeInspection     & This workbook contains the results of bridge safety   inspections conducted in 2021.                                                    \\
CampaignROI          & This workbook analyzes the return on investment (ROI) of   various marketing campaigns.                                                 \\
CourseEnrollments    & This workbook tracks student enrollment numbers for various   courses during the academic year.                                         \\
TaxFilingSummary     & This workbook summarizes the company's tax filings for the   past three years.                                                          \\
MeetingMinutes       & This workbook records the minutes and action items from weekly   department meetings.                                                   \\
PowerGridStatus      & This workbook monitors the status of the city's power grid,   including outages and repairs.                                            \\
AdBudgetAllocation   & This workbook details the allocation of the advertising budget   across different channels. 
\\ \bottomrule
\end{tabular}%
}
\end{table*}

\begin{table*}[htbp]
\centering
\caption{Description of each fine-grained operation involved in SheetRM dataset.}
\label{tab:oper_desc}
\resizebox{\textwidth}{!}{%
\begin{tabular}{lll}
\toprule
\textbf{Manipulation Category}    & \textbf{Operation}         & \textbf{Description}                                    \\ \midrule
Value Processing     & Calculate                  & Calculations and statistics.                            \\
                     & Insert                     & Insert rows or columns.                                 \\
                     & Delete                     & Delete cells, rows or columns.                          \\
                     & Auto Fill                  & Fill according to the control relationship.             \\
                     & Sort                       & Sort rows or columns in ascending or descending order.  \\
                     & Copy \& Paste              & Copy and paste cell values.                             \\
                     & Replace                    & Replace the values of a cell at a specified location.   \\
                     & Hyperlink                  & Set up hyperlinks.                                      \\
                     & Distinction                & Remove duplicates.                                      \\
                     & Filter                     & Filter specified cells according to certain conditions. \\
Worksheet Management & Create Worksheet           & Create a new worksheet.                                 \\
                     & Delete Worksheet           & Delete the specified worksheet.                         \\
                     & Rename Worksheet           & Rename the specified worksheet.                         \\
                     & Label Color                & Modify the color of worksheet name labels.              \\
                     & Page Size                  & Modify page size.                                       \\
                     & Orientation                & Set the page orientation.                               \\
Format Adjustment    & Font Name                  & Set the font category.                                  \\
                     & Font Color                 & Set the font color.                                     \\
                     & Font Size                  & Set the font size.                                      \\
                     & Bold \& Italic             & Set the font to be bold or slanted.                     \\
                     & Underline                  & Underline cell contents.                                \\
                     & Merge \& Unmerge           & Merge or split cells.                                   \\
                     & Alignment                  & Align cells horizontally or vertically.                 \\
                     & Row Height \& Column Width & Set cell row height or column width.                    \\
                     & Background Fill            & Set cell background fill color.                         \\
                     & Numeric Format             & Set cell number formatting.                             \\
Chart Design          & Chart Type                 & Set the Chart Type.                                     \\
                     & Chart Data Source          & Set the data source for the chart.                      \\
                     & Chart Caption              & Set the title of the chart.                             \\
                     & Chart Legend               & Set the Chart Legend.                                   \\
                     & Chart Position             & Specify where to place the chart.                       \\
                     & Chart Axis                 & Set the axes of a chart.                                \\
                     & Data Label                 & Set data labels for charts.                             \\
                     & Trendline                  & Add a trendline to the chart.                           \\
Content Summary       & Pivot Creation             & Create pivot table.                                     \\
                     & Summary Function           & Set statistical functions of the pivot.                 \\ \bottomrule
\end{tabular}%
}
\end{table*}

\section{Details of Code Collection for the Retriever}
\label{app:detail_retriever}
The Retriever's code comes from GitHub open-source projects and external Python libraries like \textit{openpyxl} and \textit{pandas}, focusing on high-quality, popular code to ensure data representativeness. We organized these codes by operations covered in SheetRM and then abstracted them for universality. For operations without existing code, we gathered more from the same sources or wrote code ourselves, ensuring coverage of all defined operations. The organized code is related to the corresponding tasks, mainly demonstrating the application programming interfaces and providing high-level guidance. However, the specific implementations of these APIs and the generated solutions are different. Since we abstracted and encapsulated the collected codes, we only provided information on how to operate in it, while the application of the actual data is relevant to the task scenario. Thus, we anticipate that LLMs learn from the knowledge provided by these code snippets and reflect on past trajectories to generate more robust and higher-quality solutions to the task.

\section{Dataset Details}
\label{app:dataset_detail}
The details of datasets mentioned in \cref{sec:expr_setup} are provides as follows:
\begin{itemize}[leftmargin=*]
	\item \textbf{WikiTableQuestions} includes intricate questions created by crowd workers from Wikipedia tables. These questions necessitate multiple advanced operations like comparison, aggregation, and arithmetic, demanding a detailed compositional analysis of table entries. This dataset uses CC-BY-SA-4.0 license.
	\item \textbf{FeTaQA} features free-form questions derived from tables that call for profound reasoning and comprehension. Predominantly, the questions in FetaQA arise from non-contiguous segments of the table. The performance is measured by accuracy on a test set of 2,003 samples. This dataset uses CC-BY-SA-4.0 license.
	\item \textbf{TabFact} serves as a benchmark for table-based fact verification, with crowd workers composing statements from Wikipedia tables. For instance, the statement: ``The industrial and commercial panel has four more members than the cultural and educational panel.'' must be validated as ``True'' or ``False'' based on the table information. Accuracy is reported on a smaller test set of 2,024 statements across 298 tables. It uses MIT license.
	\item \textbf{SCB} is a spreadsheet manipulation dataset, which contains 28 spreadsheets collected the Internet. The 221 spreadsheet control tasks within this dataset are generated using GPT-4, including analyzing sales data, calculating financial metrics, and visualizing data with charts. It uses GPL-3.0 license.
	\item \textbf{SheetRM} introduced in \cref{sec:benchmark} comprises of 317 real-world spreadsheet tasks. Each task includes multi-category and long-horizon manipulation sequences, along with a specific reasoning challenge. This dataset comprehensively assess the reasoning and manipulation capabilities of LLM-based agents. It also supports finer-grained and more flexible automatic evaluation. For more details of our proposed SheetRM benchmark, please refer to \cref{app:sheetrm_detail}. Our SheetRM dataset follows the CC-BY-SA-4.0 license.
\end{itemize}

\section{Implementation Details}
\label{app:imp_detail}
\textbf{Baselines.} As for table reasoning tasks, we run Binder and DATER using the official implementations. The only difference is that we revise the code to use publicly available \textit{gpt-3.5-turbo-16k-0613} as the LLM backbone instead of Codex due to its inaccessibility. We also run StructGPT on TabFact small-test set and FeTaQA using its open-sourced code with the same LLM backbone. On the proposed SheetRM, we have improved SheetCopilot based on the simplified open-source version\footnote{\url{https://github.com/BraveGroup/SheetCopilot}.} with error feedback functionality for fair comparison. For the VBA method, we adjust the prompt of SheetAgent to generate \textit{pywin32} code and remove the Retriever module due to code repository mismatch. FormaT5 and SpreadsheetCoder are implemented using the official open-sourced code. For the rest baselines, we report the performance obtained from papers. 

\noindent\textbf{LLM Backbones for SheetAgent.} In the main experiments, we select various LLMs as the backbones for our proposed SheetAgent. As for proprietary LLMs, we choose GPT-3.5, GPT-4\footnote{\url{https://platform.openai.com/docs/models}}, and Claude 3 (\texttt{claude-3-sonnet-20240229}\footnote{\url{https://docs.anthropic.com/claude/docs/models-overview}}).
In terms of open-source LLMs, we adopt Qwen-1.5 (\texttt{14b-chat}\footnote{\url{https://huggingface.co/Qwen/Qwen-14B-Chat}}) and Llama 3 (\texttt{8b-instruct}\footnote{\url{https://huggingface.co/meta-llama/Meta-Llama-3-8B-Instruct}}). Multiple versions of GPTs are involved for alignment with other baselines. Specifically, for SCB, WikiTableQuestions, FeTaQA, and TabFact, we use \texttt{gpt-3.5-turbo-16k-0613}. For our SheetRM, we employ \texttt{gpt-3.5-turbo-1106} and \texttt{gpt-4-turbo-0409}. For the embedding model in the Retriever, we adopt \texttt{text-embedding-ada-002}\footnote{See  \url{https://platform.openai.com/docs/models/embeddings}}.

\noindent\textbf{Choice of In-context Examples.} For the SCB dataset, we align with SheetCopilot by using only one in-context example. For other datasets, we utilized two in-context examples each. Specifically, for SCB, we selected the same task as SheetCopilot. We initially had VBA and SheetAgent generate a trajectory for the task under a zero-shot setting using GPT-4, then made appropriate modifications to ensure correctness. The modified trajectory was ultimately used as the in-context example. For SheetRM benchmark, we constructed two additional tasks not present in the dataset and employed VBA, SheetAgent and SheetCopilot to generate trajectories for these tasks, in the same manner, to serve as in-context examples. For other datasets, including WTQ, FeTaQA, and TabFact, where our experiments were conducted on the test sets, we chose two tasks from their respective training sets as examples. It is worth noting that our SheetAgent only uses 2 in-context examples while Binder uses 14. SheetAgent still achieves superior performance.

\noindent\textbf{Computing Power.} All the results in our experiments are obtained by running the code on a server equipped with an Intel(R) Xeon(R) Gold 6338 CPU @ 2.00GHz and 2*NVIDIA A800.

\section{Additional Experimental Results}
\label{app:additional_res}

\subsection{Full Results on Table Reasoning Tasks}
\label{app:res_reasoning_full}
Table \ref{tab:res_wtq_fact}-\ref{tab:res_feta} show the full evaluation results on table reasoning tasks.

\begin{table}[htbp]
\centering
\caption{Results of different methods on WTQ test set and TabFact small-test set. We report the accuracy metric. Best results are \textbf{bolded} and second-best results are \underline{underlined}.}
\label{tab:res_wtq_fact}
\begin{tabular}{lcc}
\toprule
\textbf{Method}                 & \textbf{WTQ}         & \textbf{TabFact}      \\ \midrule
\textbf{Fine-tuning based LLMs} & \multicolumn{1}{l}{} & \multicolumn{1}{l}{} \\
TAPAS \cite{herzig2020tapas}                           & 48.8                 & 83.9                 \\
TAPEX \cite{liu2021tapex}                          & 57.5                 & 84.2                 \\
UnifiedSKG \cite{xie2022unifiedskg}                     & 49.3                 & \textbf{85.4}                 \\
OmniTab \cite{jiang2022omnitab}                        & \underline{62.8}                 & 82.8                 \\
\textbf{Prompting based LLMs}   &                      &                      \\
GPT-3 CoT \cite{chen2023large}                      & 45.7                 & 76.0                 \\
Binder \cite{cheng2022binding}                       & 59.9                 & 82.9                 \\
DATER \cite{ye2023large}                         & 61.6           & 80.7                 \\
StructGPT \cite{jiang2023structgpt}                      & 52.2                 & 81.2        \\ \midrule
\rowcolor{lb} SheetAgent (Ours)                     & \textbf{64.4}        & \underline{84.8}           \\ \bottomrule
\end{tabular}
\end{table}

\begin{table}[htbp]
\centering
\caption{Results of different methods on FeTaQA test set. Best results are \textbf{bolded} and suboptimal results are \underline{underlined}.}
\label{tab:res_feta}
\begin{tabular}{lc}
\toprule
\textbf{Method}                 & \textbf{sacreBLEU} \\ \midrule
\textbf{Fine-tuning based LLMs} &                    \\
T5-small \cite{nan2022fetaqa}                       & 21.6               \\
T5-base \cite{nan2022fetaqa}                        & 28.1               \\
T5-large \cite{nan2022fetaqa}                       & 30.5               \\
TAPEX \cite{liu2021tapex}                         & 34.7               \\
UnifiedSKG \cite{xie2022unifiedskg}                     & 33.4               \\
PeaQA \cite{pal2022parameter}                         & 33.5               \\
OmniTab \cite{jiang2022omnitab}                         & \underline{34.9}         \\
\textbf{Prompting based LLMs}   &                    \\
GPT-3 CoT \cite{chen2023large}                      & 27.0               \\
Binder \cite{cheng2022binding}                          & 31.6               \\
DATER \cite{ye2023large}                          & 30.9               \\
StructGPT \cite{jiang2023structgpt}                      & 32.5               \\ \midrule
\rowcolor{lb} SheetAgent (Ours)                     & \textbf{36.7}      \\ \bottomrule
\end{tabular}
\end{table}

\begin{table}[htbp]
\centering
\caption{Performance comparison between SheetAgent (GPT-4) and SheetAgent (GPT-4V) on 10 representative tasks from SheetRM. Vison-enabled SheetAgent removes the Informer module.}
\label{tab:gpt4v}
\begin{tabular}{lcc}
\toprule
\textbf{Method} & \textbf{Pass@1 \up} & \textbf{SubPass@1 \up} \\ \midrule
SheetAgent (GPT-4V)                  & 40.0            & 66.5     \\
SheetAgent (GPT-4)                  & 50.0            & 74.1      \\ \bottomrule
\end{tabular}
\end{table}

\begin{figure}[htbp]
\begin{center}
\centerline{\includegraphics[width=\columnwidth]{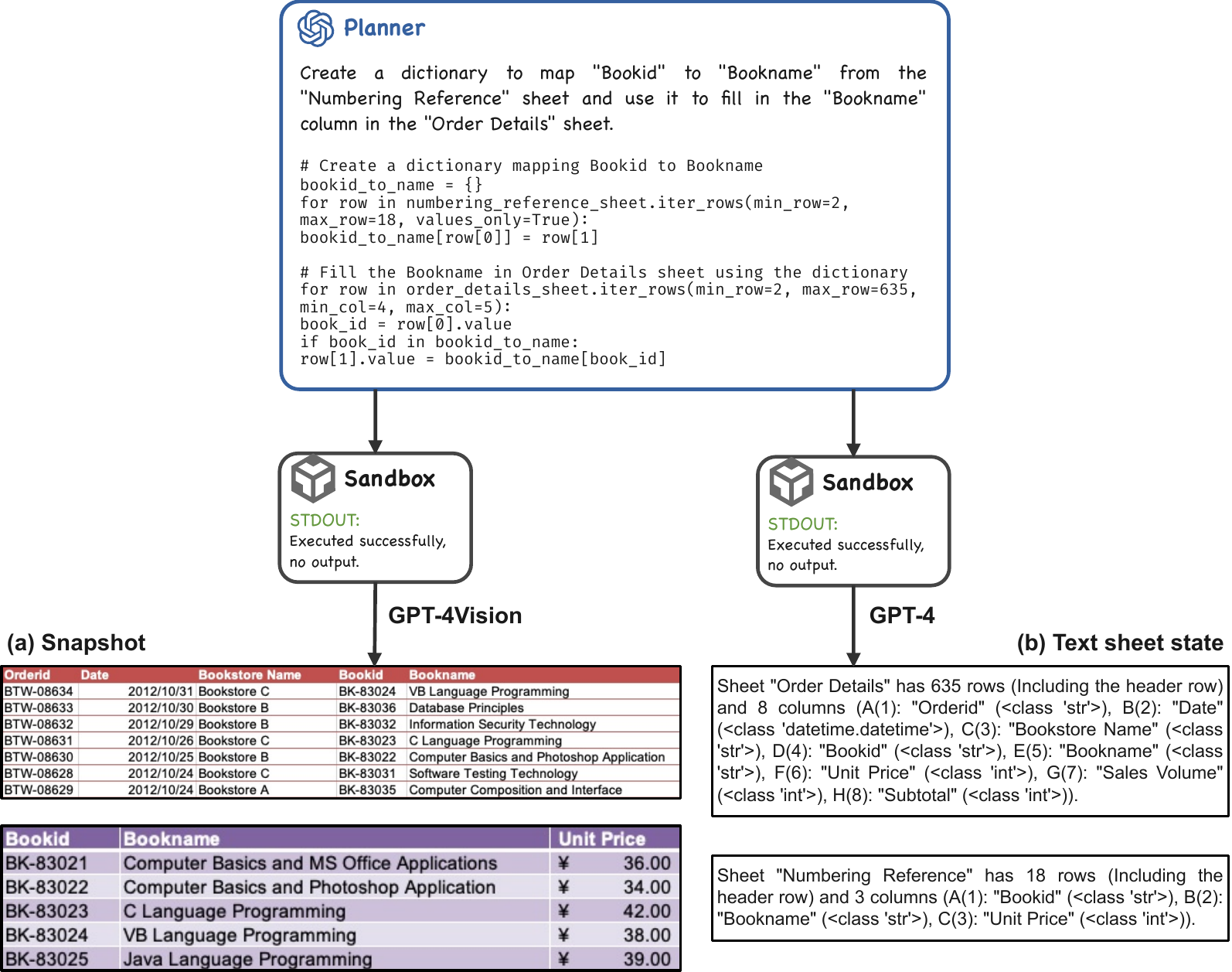}}
\caption{An illustration depicting the differences in sheet state between GPT-4V and GPT-4. For brevity, the Informer and Retriever modules are excluded. The snapshots (namely the visual representation of sheet state) are partial due to the limitation of spreadsheet scale.}
\label{fig:sa_vision}
\end{center}
\end{figure}

\subsection{Vision-Enabled SheetAgent}
We have explored the potential of leveraging GPT-4V(ision)'s visual capabilities by substituting spreadsheet snapshots for the text-modal sheet state in the observation. Given the cost of GPT-4V and the challenges in automatic snapshot capture of spreadsheets, we test this approach with 10 representative tasks from the SheetRM dataset. We have ensured these tasks span all five manipulation categories defined in SheetRM. As vision-eanbled SheetAgent can observe full state of spreadsheets, we remove the Informer module for fair comparison. Results are presented in \cref{tab:gpt4v}. Through this intriguing experiment, we observe that when tasks involved visual elements, such as formatting and chart modification, SheetAgent can better adjust styles based on the visual feedback, improving task completion. However, SheetAgent (GPT-4V) has difficulty processing large-scale spreadsheets and correctly identifying sheet data due to low image solution, resulting in task failure. We also obtain an interesting finding that there exists an overlap between the visual aspect of GPT-4V and the Informer module. While GPT-4V allows SheetAgent to perceive multimodal content within spreadsheets (charts, pivot tables, frozen panes, etc.), it faces challenges capturing accurate information in larger-scale tables compared with the Informer. We leave this for our future work.

\subsection{Ablation Study on Table Representation}
\label{app:table_repr}
\begin{figure}[htbp]
\centering
\includegraphics[width=\columnwidth]{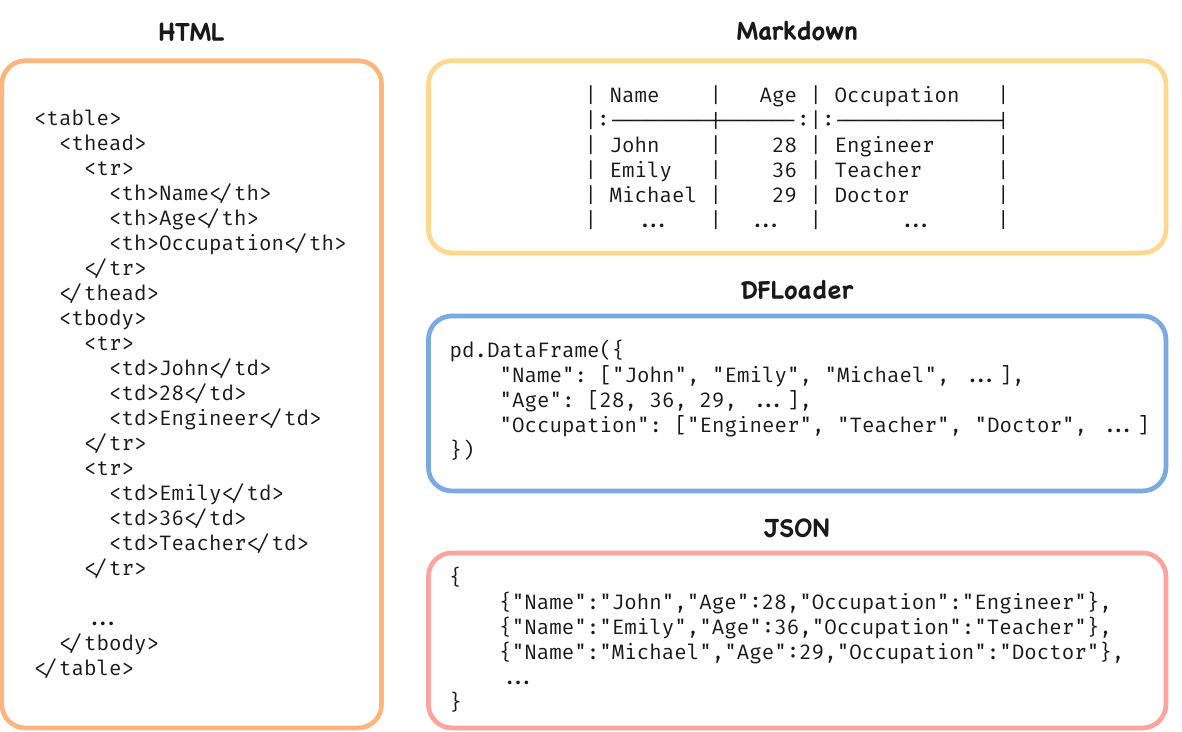}
\caption{An illustration of four different table representations we use in our experiments.}
\label{fig:table_repr}
\end{figure}

Tabular data is a kind of information-dense structured data, it is crucial to design reliable representations to enable LLMs clearly recognize the attribute relationships. To investigate what representation can better help LLMs to reason over tables. We ablate 4 prevalent table representations: JSON, DFLoader, Markdown, and HTML for SheetAgent on WikiTableQuestions and SheetRM. We provide an illustration of these representations, as shown in \cref{fig:table_repr}. Notably, DFLoader is represented by the corresponding Python code snippet that uses the \textit{pandas} DataFrame API to define the table. The results shown in \cref{tab:table_repr} reveal that JSON outperform other formats. HTML format achieves a suboptimal result on WTQ, but ranks lowest on SheetRM. Its open-and-close structure helps LLMs understand better, but the verbosity risks exceeding token limits, thus hindering efficiency. We also observe that DFLoader format achieve commendable results, possibly due to its code structure, which might be easier for LLMs to comprehend. Overall, JSON is a preferable choice for both reasoning intensive tasks, like WTQ, and long-horizon tasks with fewer reasoning elements, such as SheetRM.

\subsection{Ablation Study on LLM Temperature}
We conduct evaluations of our method using the proposed SheetRM dataset under varying conditions by adjusting the temperature settings to investigate the impact of temperature on the performance of LLMs. For these experiments, \texttt{gpt-3.5-turbo-1106} is selected as the LLM backbone. Our findings reveal that our method, SheetAgent, achieves its best performance at a temperature of 0.0, with minor performance fluctuations observed at a temperature of 0.2. However, a noticeable decline in performance across all metrics occurs when the temperature is increased to 0.4. This trend suggests that higher temperature settings lead to more unpredictable outcomes from SheetAgent, reflecting a decrease in the stability and reliability of the solutions it generates.

\begin{table}[!t]
\centering
\caption{Ablation study on the temperature of LLM.}
\label{tab:abla_temp}
\begin{tabular}{cccc}
\toprule
\textbf{Temperature} & \textbf{Exec@1 \up} & \textbf{Pass@1 \up} & \textbf{SubPass@1 \up} \\ \midrule
0.0                  & 92.4            & 31.2            & 69.8               \\
0.2                  & 93.5            & 30.3            & 68.1               \\
0.4                  & 90.2            & 28.7            & 66.2               \\ \bottomrule
\end{tabular}%
\end{table}

\begin{table}[!t]
\centering
\caption{Token and time consumption comparison. Consumption of tokens is calculated by stage.}
\label{tab:cost}
\resizebox{\columnwidth}{!}{%
\begin{tabular}{lp{0.65\columnwidth}c}
\toprule
\textbf{Method}        & \multicolumn{1}{c}{\textbf{Avg. \# Tokens}}                                                  & \textbf{Avg. Time (s)} \\ \midrule
SheetAgent (GPT-3.5)   & System Prompt: 324 + Few-shot Demonstrations: 2013 + Planner: 589.7 + Informer: 513.9 + Retriever: 625.3 = \textbf{4065.9} & 6.9                    \\
SheetCopilot (GPT-3.5) & System Prompt: 1895 + Few-shot Demonstrations: 1592 +  Plan: 628.5 = 4115.5                                          & \textbf{5.8}                    \\ \bottomrule
\end{tabular}
}
\end{table}

\begin{figure}[!t]
\centering
\includegraphics[width=\columnwidth]{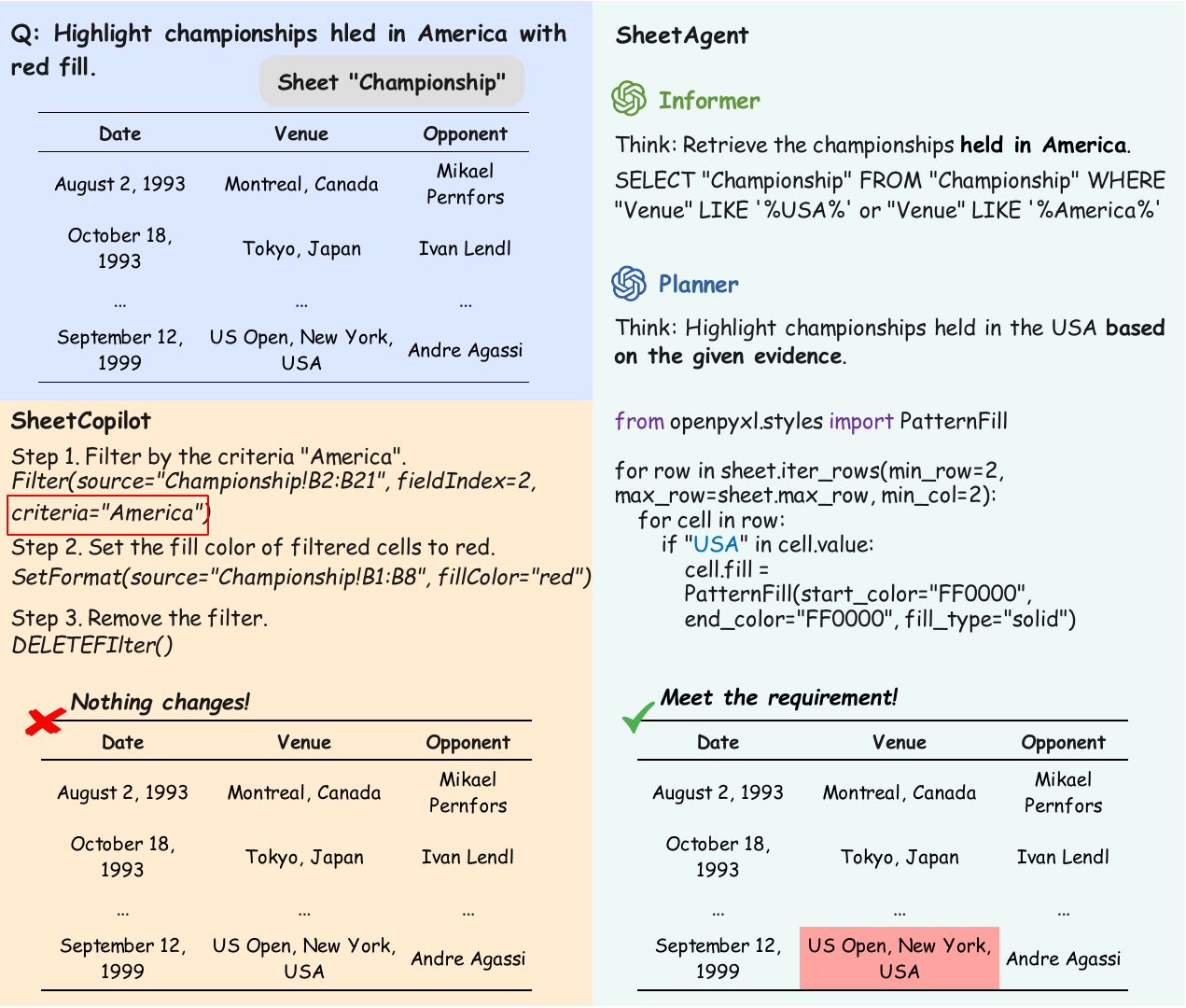}
\caption{A comparison between SheetAgent and SheetCopilot on a spreadsheet task with reasoning challenges. SheetCopilot generates a rigid solution that fails to fulfill the instruction. SheetAgent identifies the task intention and gives a correct solution.}
\label{fig:case_compare}
\end{figure}

\subsection{Performance-Cost Analysis}
We have conducted extra experiments to calculate the token and time consumption of our method on SheetRM. We use GPT-3.5 as the LLM backbone. We compare our proposed SheetAgent with SheetCopilot. It is noteworthy that since SheetCopilot is insufficiently capable of fulfilling a complete task from SheetRM, we select 20 subtasks for which both can generate successful trajectories and calculate metrics based on these. The results are presented in \cref{tab:cost}. On average, our approach consumes fewer tokens compared to SheetCopilot, primarily because SheetCopilot often makes errors, which leads to reflection. However, SheetAgent involves querying multiple LLMs and a vector database, which places us at a disadvantage in terms of time efficiency.

As for cost, we have calculated the cost of successful trajectories by our SheetAgent (GPT-3.5) on SheetRM. The average cost to finish a complete task is \$0.0049. Notably, excellent results of SheetAgent shown in \cref{tab:res_scb_sheetrm} can be achieved even with relatively cheap backbone GPT-3.5 and Claude 3 Sonnet, which is a trade-off between cost and performance. We believe the superior performance of SheetAgent compared to other methods justify this resource use.

\section{An Illustrative Case Between SheetAgent and SheetCopilot}
\label{app:case_compare}
\cref{fig:case_compare} presents a case with reasoning challenge. For SheetAgent (GPT-3.5), the Informer accurately selects the key evidence related to the task instruction. The Planner correctly fulfill the task based on the evidence. In contrast, SheetCopilot (GPT-4) merely offers a rigid solution that fails to complete the task despite its successful execution.

\begin{figure}[!t]
\centering
\includegraphics[width=\columnwidth]{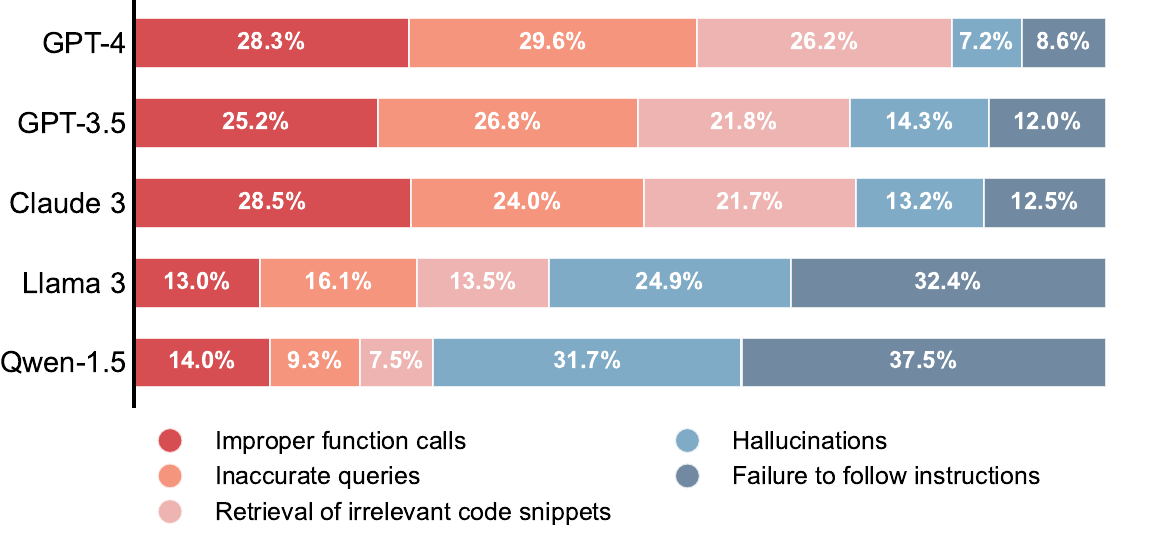}
\caption{Distributions of different error cases for various LLM backbones on the SheetRM dataset.}
\label{fig:error}
\end{figure}

\section{Failure Cases Analysis}
\label{app:failure_case}
To gain a clearer understanding of the differences between the LLM backbones compared in Sections \cref{sec:rq1} and \cref{sec:rq2}, we conduct a detailed error analysis to determine the causes and locations of failures. We classify the reasons for failure as follows:
\begin{itemize}[leftmargin=*]
\item \textbf{Improper function calls:} The Planner inaccurately invokes function interfaces for certain operations. For example, it uses \texttt{chart.set\_title("Chart Title")} instead of \texttt{chart.title = "Chart Title"}, resulting in an \texttt{AttributeError}. Additionally, it performs \texttt{worksheet.cell\_range()} (a method deprecated in newer versions of \textit{openpyxl}) instead of \texttt{worksheet.iter\_rows()} or \texttt{worksheet.iter\_cols()}.
\item \textbf{Inaccurate queries:} The Informer generates imprecise SQL statements leading to incorrect or incomplete information being input. For example, it attempts to query books related to office software but produces a statement like \texttt{SELECT * FROM w WHERE BookName LIKE `\%Excel\%' OR BookName LIKE `\%PowerPoint\%'} while ignoring \texttt{Word}.
\item \textbf{Retrieval of irrelevant code snippets:} The Retriever sources irrelevant code fragments, which impedes the Planner's correction process. This happens due to similarities in code segments within the code repository, resulting in the retrieval of irrelevant code.
\item \textbf{Hallucinations:} It performs operations on rows and columns that are out of scope, ignoring the actual data, or creates data that does not exist.
\item \textbf{Failure to follow instructions:} It terminates tasks prematurely or fails to comply with the given instructions. For example, it only completes a part of subtasks from a long-horizon task or highlights entries in colors not specified by the task.
\end{itemize}

We present the proportions of different failure cases for various LLM backbones on our SheetRM in \cref{fig:error}. Combining the results in \cref{fig:res_backbone_sheetrm}, we can observe that: (\rmnum{1}) LLMs with poorer performance on the benchmark (e.g., llama3-8b-instruct and qwen-14b-chat) have a significantly higher proportion of errors related to ``hallucinations'' and ``failure to follow instructions''. In contrast, well-tuned LLMs with extensive parameters, such as GPT-3.5, GPT-4, and Claude 3 Sonnet, have their errors predominantly concentrated in ``improper function calls'', ``inaccurate queries'', and ``retrieval of irrelevant code snippets''. This indicates that advanced LLMs perform better on complex tasks, whereas smaller open-sourced LLMs struggle significantly with understanding and executing instructions. (\rmnum{2}) Among proprietary LLMs, GPT-3.5 and GPT-4 exhibit similar error distributions, with high proportions of errors in ``inaccurate queries'' and ``improper function calls'' (29.6\% and 28.3\% for GPT-4, 26.8\% and 25.2\% for GPT-3.5, respectively). In contrast, Claude 3 Sonnet shows a different pattern, with a similar proportion of ``inaccurate queries'' (24.0\%) but a relatively higher proportion of ``improper function calls'' (28.5\%). This may reflect that GPTs are adept at generating proficient Python code, while Claude can better understand complicated instructions and translate them into accurate SQLs. (\rmnum{3}) Smaller open-source LLMs, such as llama3-8b-instruct and qwen-14b-chat, display similar error patterns, primarily in ``hallucinations'' and ``failure to follow instructions''. Llama3-8b-instruct possesses a ``failure to follow instructions'' error rate of 32.4\%, whereas qwen-14b-chat has a significantly higher rate of 37.5\%. This phenomenon may be attributed to their training corpus and model scale.

We further perform a deep analysis of specific failure cases across different LLM backbones, which reveals distinct patterns and challenges. GPT-4 and GPT-3.5 are prone to make errors in generating correct SQLs. After inspecting the specific bad cases, we find that in most cases, they understand the task instruction but generate SQLs that semantically fail to fulfill the task requirements. In other cases, they generate syntactically incorrect SQL statements that cause execution to fail. 
Differently, Claude 3 Sonnet owns a highest rate of improper function calls but fewer errors of inaccurate queries. It usually calls a function that does not exist or is deprecated, or misunderstands the function usage. For instance, it uses \textit{openpyxl}'s \texttt{iter\_rows()} function to iterate through the spreadsheet. The exact code it produces is \texttt{for row in ws.iter\_rows(min\_row=1, max\_row=10, max\_col="E"):}, where \texttt{max\_col} should be an integer instead of a string. 
Llama3-8b-instruct and qwen-14b-chat share the highest proportions of instruction-following failures and hallucinations, suggesting difficulties in maintaining task context and adhering to long-horizon instructions. We note that there are a large number of incomplete solutions in the llama3-8b-instruct failure case, due in large part to its limited context length of 8K. For qwen-14b-chat, we observe that it can hardly follow the complicated and long-horizon task instructions, and tends to generate irrelevant contents. We assume this may have something to do with its training strategy and corpus.

We have further proposed potential strategies to overcome the proposed failure cases, which may provide insights for future research in this community:
\textbf{Regarding addressing improper function calls,} we found conflicts between the LLM's training corpus on \textit{openpyxl} versions and current versions. Enhancing understandings of library functions through fine-tuning or tool augmentation might mitigate this.
\textbf{For inaccurate queries,} improving model training with diverse SQL examples through fine-tuning and incorporating a validation layer to check queries against database schemas could enhance accuracy.
\textbf{To combat irrelevant code snippet retrieval,} refining the code repository with detailed descriptions of each example's functionality and intended task scenarios could improve retrieval accuracy.
\textbf{For hallucinations and failure to follow instructions,} we attribute these to the model's inherent limitations, noticing a significant increase in these issues on weaker LLM backbones like llama3-8b-instruct and qwen-14b-chat. Switching to a more robust LLM might alleviate these problems. Explicitly managing task progress (e.g., adding a task decomposition module for procedural execution) or incorporating an LLM-driven Critic module (for sanity check on generated solutions) could also partially address these issues.

\section{Prompts}
\label{app:prompt}
\begin{figure*}[htbp]
\centering
\includegraphics[width=0.9\textwidth]{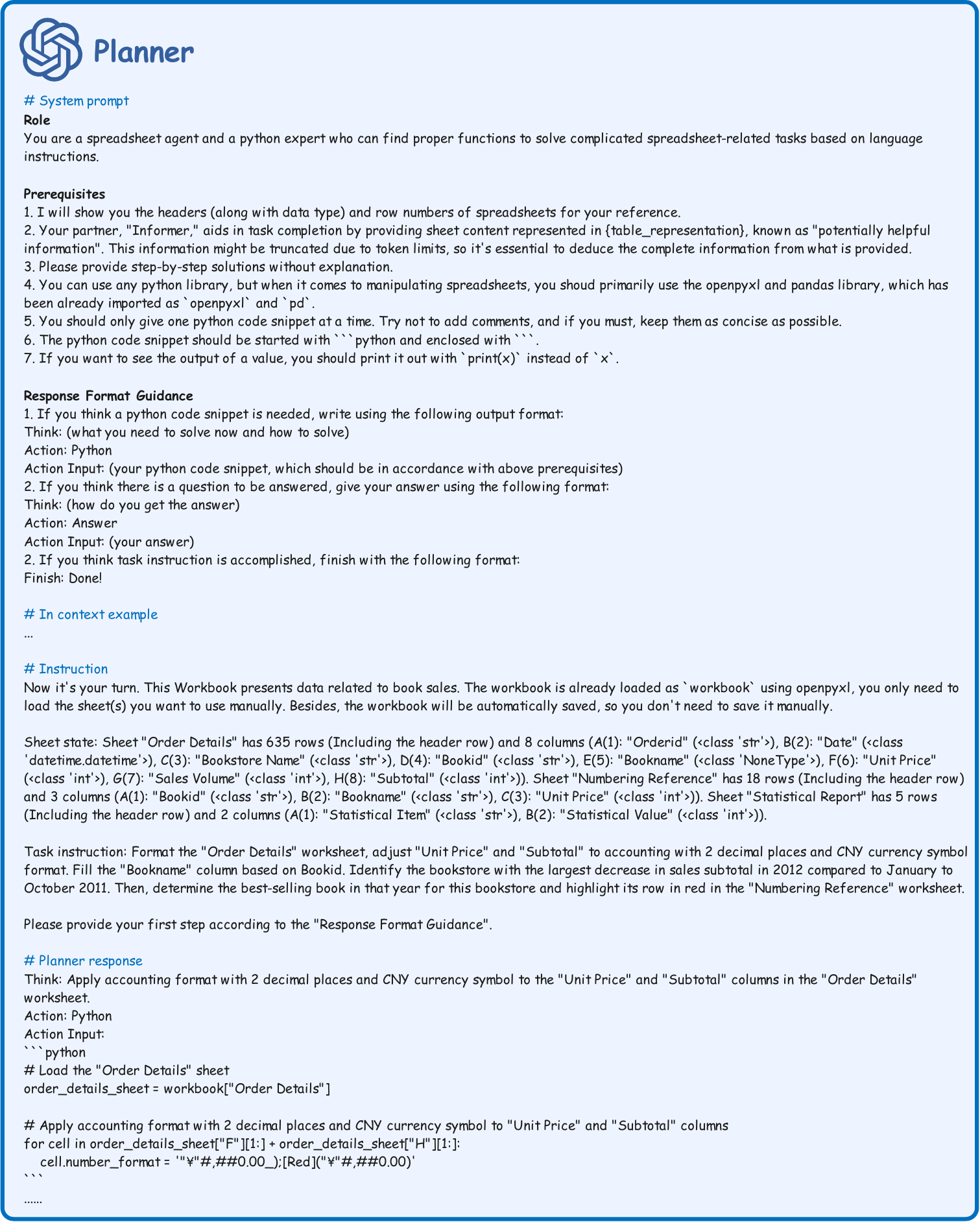}
\caption{A prompt template for the Planner.}
\label{fig:prompt_planner}
\end{figure*}

\begin{figure*}[htbp]
\centering
\includegraphics[width=0.9\textwidth]{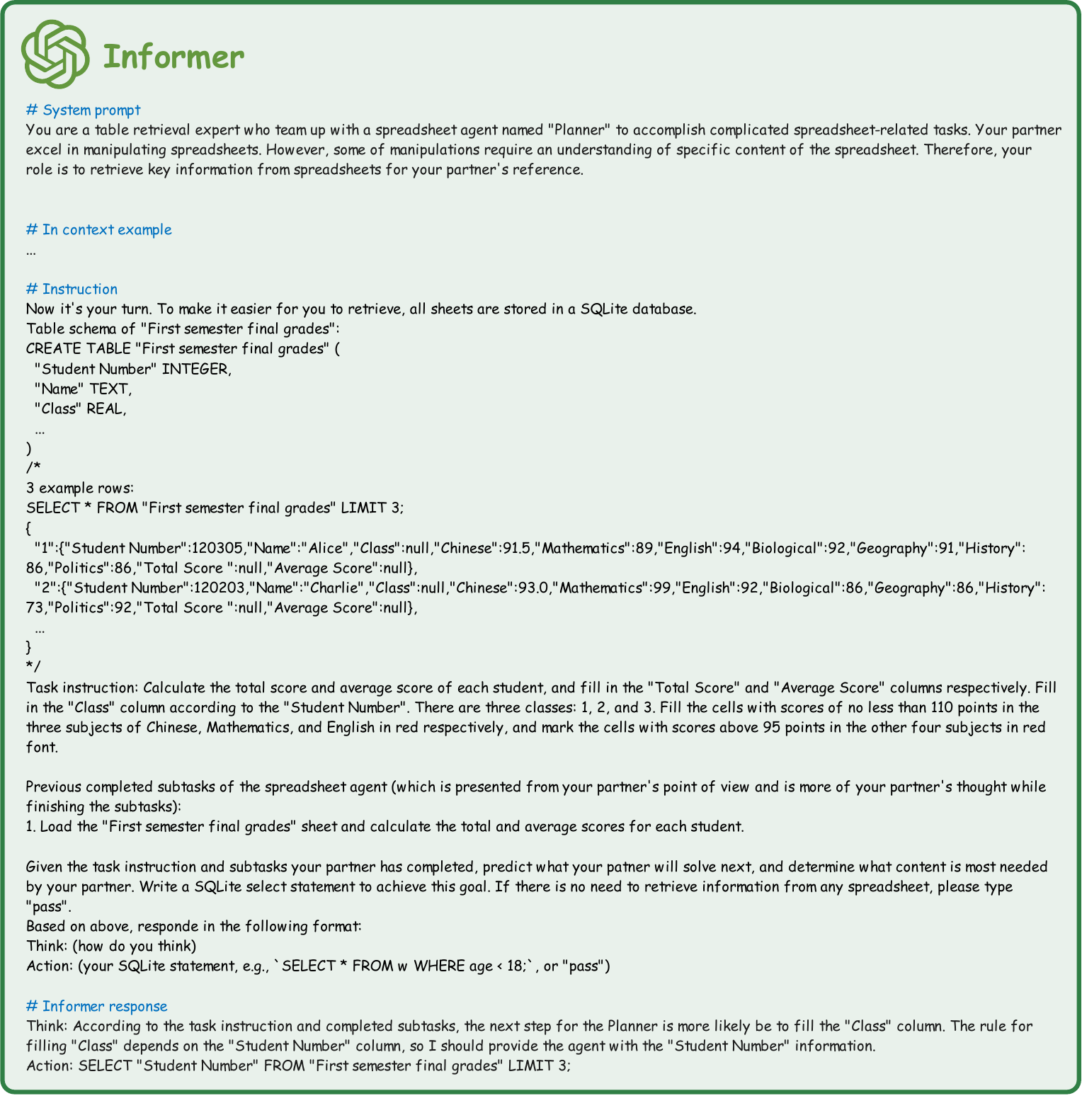}
\caption{A prompt template for the Informer.}
\label{fig:prompt_informer}
\end{figure*}

We provide the prompts for Planner and Informer, as shown in \cref{fig:prompt_planner} and \cref{fig:prompt_informer}. Additional prompts for subtask generation can be found in our \href{https://sheetagent.github.io/}{project website}.

\section{Limitations and Potential Social Impact}
\label{app:limits}
We list the limitations of our proposed SheetAgent as follows:
\begin{itemize}[leftmargin=*]
\item \textbf{Library limitations:} SheetAgent automates spreadsheet tasks through Python code generation, utilizing libraries like \textit{openpyxl} and \textit{pandas}. Although this has covered a wide range of operations, it is still missing some customizable functionality at the software level. For instance, complex spreadsheet manipulations that involve advanced Excel features such as pivot tables, macros, or specific formatting options are not fully supported. Enhancements in library capabilities or integration with additional tools could address these gaps.
\item \textbf{High token usage:} Like existing research to automate spreadsheet manipulation \cite{li2024sheetcopilot}, SheetAgent inevitably faces higher token usage for long-horizon tasks. This can lead to increased computational costs and slower processing times. Future work will focus on optimizing task descriptions through more efficient prompting techniques or manual refinement to reduce token consumption and improve overall efficiency.
\end{itemize}

The implementation of SheetAgent has the potential to bring about several positive social impacts. By automating repetitive and time-consuming spreadsheet tasks, SheetAgent can significantly enhance productivity and efficiency in various industries. This can free up human resources for more strategic and creative work, ultimately leading to better utilization of talent and skills. Additionally, SheetAgent can democratize access to advanced data analysis and processing, making these capabilities available to a broader audience, including individuals with limited technical expertise. This democratization can empower more people to leverage data for informed decision-making and innovation.

However, the introduction of SheetAgent might pose negative social impacts. As with any automation technology, there is a risk of job displacement for roles traditionally centered around manual spreadsheet manipulation. This could lead to economic and social challenges for affected individuals. Moreover, the reliance on computational resources for running SheetAgent, especially for large-scale or long-horizon tasks, could contribute to environmental concerns such as increased energy consumption. Addressing these issues requires proactive measures, including reskilling and upskilling programs to help displaced workers transition to new roles and optimizing the efficiency of SheetAgent to minimize its environmental footprint. Ethical considerations must also be prioritized to ensure transparency, fairness, and the safeguarding of user data privacy and security.

\end{document}